\documentclass[10pt,journal,compsoc]{IEEEtran}
\usepackage{times}
\usepackage{epsfig}
\usepackage{graphicx}
\usepackage{amsmath}
\usepackage{amssymb}
\usepackage{epsfig,subfigure,epstopdf,multirow}
\usepackage{booktabs,array}
\usepackage[pagebackref=true,breaklinks=true,letterpaper=true,colorlinks,bookmarks=false]{hyperref}
\ifCLASSOPTIONcompsoc
  \usepackage[nocompress]{cite}
\else
  \usepackage{cite}
\fi
\ifCLASSINFOpdf
\else
\fi
\begin{document}
%
\title{Grid Anchor based Image Cropping: A New Benchmark and An Efficient Model}
%
%
%
%

\author{Hui~Zeng,
        Lida~Li,
        Zisheng~Cao,
        and~Lei~Zhang,~\IEEEmembership{Fellow,~IEEE}
\IEEEcompsocitemizethanks{\IEEEcompsocthanksitem H. Zeng, L. Li and L. Zhang are with the Department of Computing, The
Hong Kong Polytechnic University, Hong Kong. (email: \{cshzeng, cslli, cslzhang\}@comp.polyu.edu.hk).
\IEEEcompsocthanksitem Z. Cao is with Camera Group of DJI Innovations Co., Ltd, Shenzhen, China (e-mail: zisheng.cao@dji.com).}}
\IEEEtitleabstractindextext{%
\begin{abstract}
Image cropping aims to improve the composition as well as aesthetic quality of an image by removing extraneous content from it. Most of the existing image cropping databases provide only one or several human-annotated bounding boxes as the groundtruths, which can hardly reflect the non-uniqueness and flexibility of image cropping in practice. The employed evaluation metrics such as intersection-over-union cannot reliably reflect the real performance of a cropping model, either. This work revisits the problem of image cropping, and presents a grid anchor based formulation by considering the special properties and requirements (e.g., local redundancy, content preservation, aspect ratio) of image cropping. Our formulation reduces the searching space of candidate crops from millions to no more than ninety. Consequently, a grid anchor based cropping benchmark is constructed, where all crops of each image are annotated and more reliable evaluation metrics are defined. To meet the practical demands of robust performance and high efficiency, we also design an effective and lightweight cropping model. By simultaneously considering the region of interest and region of discard, and leveraging multi-scale information, our model can robustly output visually pleasing crops for images of different scenes. With less than 2.5M parameters, our model runs at a speed of 200 FPS on one single GTX 1080Ti GPU and 12 FPS on one i7-6800K CPU. The code is available at: \url{https://github.com/HuiZeng/Grid-Anchor-based-Image-Cropping-Pytorch}.
\end{abstract}

\begin{IEEEkeywords}
Image cropping, photo cropping, image aesthetics, deep learning.
\end{IEEEkeywords}}

\maketitle

\IEEEdisplaynontitleabstractindextext

%
\IEEEpeerreviewmaketitle

\IEEEraisesectionheading{\section{Introduction}\label{sec:introduction}}

Cropping is an important and widely used operation to improve the aesthetic quality of captured images. It aims to remove the extraneous contents of an image, change its aspect ratio and consequently improve its composition \cite{wiki:xxx}. Cropping is a high-frequency need in photography but it is tedious when a large number of images are to be cropped. Therefore, automatic image cropping has been attracting much interest in both academia and industry in past decades \cite{chen2003visual,chor2006system,jogo2007image,yan2013learning,fang2014automatic,downing2015automated,bhatt2015multifunctional,
chen2016automatic,chen2017quantitative,wang2017deep,chedeau2017image,li2018a2}.

Early researches on image cropping mostly focused on cropping the major subject or important region of an image for small displays \cite{chen2003visual,ciocca2007self} or generating image thumbnails \cite{suh2003automatic,marchesotti2009framework}. Attention scores or saliency values were the major considerations of these methods \cite{santella2006gaze,stentiford2007attention}. With little consideration on the overall image composition, the attention-based methods may lead to visually unpleasing outputs \cite{yan2013learning}. Moreover, user study was employed as the major criteria to subjectively evaluate cropping performance, making it very difficult to objectively compare different methods.

Recently, several benchmark databases have been released for image cropping research \cite{yan2013learning,fang2014automatic,chen2017quantitative}. On these databases, one or several bounding boxes were annotated by experienced human subjects as ``groundtruth" crops for each image. Two objective metrics, namely intersection-over-union (IoU) and boundary displacement error (BDE) \cite{freixenet2002yet}, were defined to evaluate the performance of image cropping models on these databases. These public benchmarks enable many researchers to develop and test their cropping models, significantly facilitating the research on automatic image cropping \cite{yan2013learning,deng2017image,wang2017deep,chen2017quantitative,chen2017learning,deng2017aesthetic,guo2017automatic,li2018a2,wei2018good}.

\begin{figure}[t]
\centering
\subfigure{
\begin{minipage}[b]{1.0\linewidth}
\includegraphics[width=1.0\textwidth]{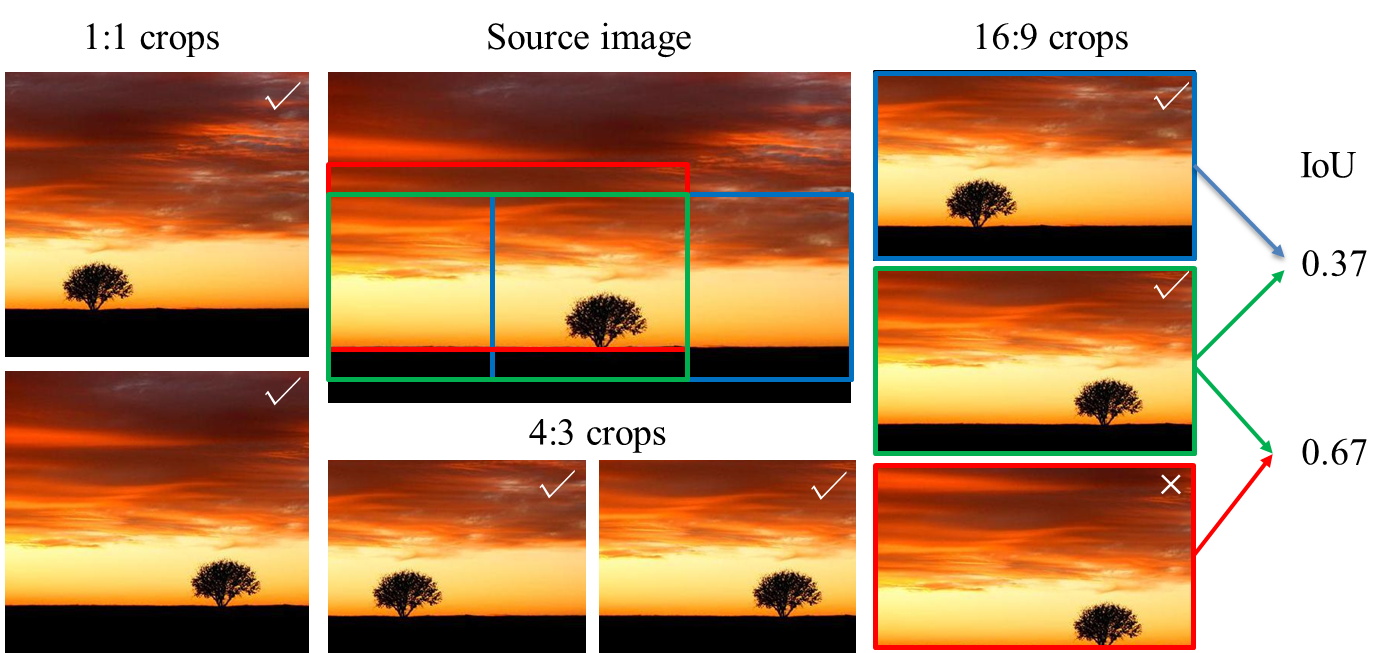}
\end{minipage}}%
\caption{The property of non-uniqueness of image cropping. Given a source image, many good crops (labeled with ``$\surd$") can be obtained under different aspect ratios (e.g., 1:1, 4:3, 16:9). Even under the same aspect ratio, there are still multiple acceptable crops. Regarding the three crops with 16:9 aspect ratio, by taking the middle one as the groundtruth, the bottom one (a bad crop, labeled with ``$\times$") will have obviously larger IoU (intersection-over-union) than the top one but with worse aesthetic quality. This shows that IoU is not a reliable metric to evaluate cropping quality.}
\label{figure:problems}
\end{figure}

Though many efforts have been made, there are several intractable challenges caused by the special properties of image cropping.
As illustrated in Fig. \ref{figure:problems}, image cropping is naturally a subjective and flexible task without unique solution. Good crops can vary significantly under different requirements of aspect ratio and/or resolution. Even under certain aspect ratio or resolution constraint, acceptable crops can also vary.  Such a high degree of freedom makes the existing cropping databases, which have only one or several annotations, difficult to learn reliable and robust cropping models.

The commonly employed IoU and BDE metrics are unreliable to evaluate the performance of image cropping models either. Referring to the three crops with 16:9 aspect ratio in Fig. \ref{figure:problems}, by taking the middle one as the groundtruth, the bottom one, which is a bad crop, has obviously larger IoU than the top one, which is a good crop. Such a problem can be more clearly observed from Table \ref{table:Performance comparison}. By using IoU to evaluate the performance of recent works \cite{yan2013learning,wang2017deep,chen2017quantitative,chen2017learning,li2018a2} on the existing cropping benchmarks ICDB \cite{yan2013learning} and FCDB \cite{chen2017quantitative}, most of them have even worse performance than the two simplest baselines: no cropping (i.e., take the source image as cropping output, denoted by Baseline\_N) and central crop (i.e., crop the central part whose width and height are 0.9 time of the source image, denoted by Baseline\_C).

\begin{table}[t]
\footnotesize
\centering
\caption{IoU scores of recent representative works and the developed models in this work on two existing cropping benchmarks in comparison with two simplest baselines. Baseline\_N simply calculates the IoU between the groundtruth and source image without cropping. Baseline\_C crops the central part whose width and height are 0.9 time of the source image.}
\label{table:Performance comparison}
\begin{tabular}{|c|ccc|c|}
\hline
\multirow{2}{*}{Method} & \multicolumn{3}{c|}{ICDB\cite{yan2013learning}} & \multirow{2}{*}{FCDB\cite{chen2017quantitative}}               \\\cline{2-4}
                                            & Set 1 & Set 2 & Set 3    &                   \\\hline
Yan \textit{et al.} \cite{yan2013learning}            & 0.7487 & 0.7288 & 0.7322 & --  \\
Chen \textit{et al.} \cite{chen2017quantitative}      & 0.6683 & 0.6618 & 0.6483 & 0.6020  \\
Chen \textit{et al.} \cite{chen2017learning}          & 0.7640 & 0.7529 & 0.7333 & \textbf{0.6802}  \\
Wang \textit{et al.} \cite{wang2017deep}              & 0.8130  & 0.8060  & 0.8160 & 0.6500  \\
Wang \textit{et al.} \cite{wang2018deep}              & 0.8150  & 0.8100  & \textbf{0.8300} & --  \\
Li \textit{et al.} \cite{li2018a2}                    & 0.8019 & 0.7961 & 0.7902 & 0.6633  \\\hline\hline
Baseline\_N                                 & \textbf{0.8237} & \textbf{0.8299} & 0.8079 & 0.6379  \\
Baseline\_C                                 & 0.7843 & 0.7599 & 0.7636 & 0.6647  \\\hline\hline
GAIC (Mobile-V2)                            & 0.8179 & 0.8150 & 0.8070 & 0.6735  \\
GAIC (Shuffle-V2)                           & 0.8199 & 0.8170 & 0.8050 & 0.6751  \\\hline
\end{tabular}
\end{table}

The special properties of image cropping make it a challenging task to train an effective and efficient cropping model. On one hand, since the annotation of image cropping (which requires good knowledge and experience in photography) is very expensive \cite{chen2017quantitative}, existing cropping databases \cite{yan2013learning,fang2014automatic,chen2017quantitative} provide only one or several annotated crops for about 1,000 source images. On the other hand, the searching space of image cropping is very huge, with millions of candidate crops for each image. Clearly, the amount of annotated data in current databases is insufficient to train a robust cropping model. The unreliable evaluation metrics further constrain the research progress on this topic.

In order to address the above issues, we reconsider the problem of image cropping and propose a new approach, namely grid anchor based image cropping, to accomplish this challenging task in a reliable and efficient manner. Our contributions are threefold.
\begin{itemize}
\item We propose a grid anchor based formulation for image cropping by considering the special properties and requirements (e.g., local redundancy, content preservation, aspect ratio) of this problem. Our formulation reduces the number of candidate crops from millions to no more than ninety, providing an effective solution to satisfy the practical requirements of image cropping.
\item Based on our formulation, we construct a new image cropping database with exhaustive annotations for each source image. With a total of 106,860 annotated candidate crops and each crop annotated by 7 experienced human subjects, our database provides a good platform to learn robust image cropping models. We also define three new types of metrics which can more reliably evaluate the performance of learned cropping models than the IoU and BDE used in previous datasets.
\item We design an effective and efficient image cropping model under the convolutional neural network (CNN) architecture. Specifically, our model first extracts multi-scale features from the input image and then models both the region of interest and region of discard to stably output a visually pleasing crop. Leveraging the recent advances in designing efficient CNN models \cite{sandler2018MobileNetV2,ma2018shufflenet}, our cropping model contains less than 2.5M parameters, and runs at a speed of up to 200 FPS on one single GTX 1080Ti and 12 FPS on CPU.
\end{itemize}

This paper extends our conference version \cite{zeng2019reliable} in four aspects. (1) More evaluation metrics are defined to evaluate more comprehensively the cropping models. (2) The feature extraction modules (VGG16 \cite{simonyan2014very} and ResNet50 \cite{he2016deep}) in the conference version are replaced by more efficient architectures (MobileNetV2 \cite{sandler2018MobileNetV2} and ShuffleNetV2 \cite{ma2018shufflenet}), which significantly improve the efficiency of our cropping model without sacrificing the performance. (3) A multi-scale feature extraction architecture is designed to more effectively handle the images with varying scales of objects. (4) A set of effective data augmentation strategies are employed for learning photo composition, which further improve the performance of trained model. With all these improvements, our new model has much smaller size, much higher efficiency and much better cropping results.

\section{Related work}

In this section, we summarize the existing image cropping datasets and evaluation metrics, representative image cropping methods and efforts made on improving cropping efficiency.

\subsection{Image cropping datasets and evaluation metrics}

Although the research of image cropping has been lasting for more than one decade, subjective assessment was employed as the major evaluation criteria for a long time because of the highly subjective nature and the expensive annotation cost of image cropping. Yan \textit{et al.} \cite{yan2013learning} constructed the first cropping dataset, which consists of 950 images. Each image was manually cropped by three photographers. Contemporarily, Feng \textit{et al.} \cite{fang2014automatic} constructed a similar cropping dataset which contains 500 images with each image cropped by 10 expert users. Both datasets employed the IoU and BDE to evaluate the cropping performance. Unfortunately, the limited number of annotated crops is insufficient to learn a robust cropping model and the evaluation metrics are unreliable for performance evaluation. To obtain more annotations, Chen \textit{et al.} \cite{chen2017quantitative} proposed a pairwise annotation strategy. They built a cropping dataset consisting of 1,743 images and 31,430 annotated pairs of subviews. Using a two-stage annotation protocol, Wei \textit{et al.} \cite{wei2018good} constructed a large scale comparative photo composition (CPC) database which can generate more than 1 million view pairs. Although the pairwise annotation strategy provides an efficient way to collect more training samples, the candidate crops in both datasets are either randomly generated or randomly selected from cropping results generated by previous cropping methods. These candidate crops are unable to provide more reliable and effective evaluation metrics for image cropping.

Different from the previous ones, our dataset is constructed under a new formulation of image cropping. Our dense annotations not only provide extensive information for training cropping model but also enable us to define new evaluation metrics to more reliably evaluate the cropping performance.

\subsection{Image cropping methods}

The existing image cropping methods can be divided into three categories according to their major drives.

\textbf{Attention-driven methods.} Earlier methods are mostly attention-driven, aiming to identify the major subject or the most informative region of an image. Most of them \cite{chen2003visual,suh2003automatic,stentiford2007attention,marchesotti2009framework} resort to a saliency detection algorithm (e.g. \cite{itti1998model}) to get an attention map of an image, and search for a cropping window with the highest attention value. Some methods also employ face detection \cite{zhang2005auto} or gaze interaction \cite{santella2006gaze} to find the important region of an image. However, a crop with high attention value may not necessarily be aesthetically pleasing.

\textbf{Aesthetic-driven methods.} The aesthetic-driven methods improve the attention-based methods by emphasizing the overall aesthetic quality of images. These methods \cite{zhang2005auto,nishiyama2009sensation,cheng2010learning,liu2010optimizing,yan2013learning,zhang2013probabilistic,fang2014automatic,zhang2014weakly} usually design a set of hand-crafted features to characterize the image aesthetic properties or composition rules. Some methods  further deign quality measures \cite{zhang2005auto,liu2010optimizing} to evaluate the quality of candidate crops, while some resort to training an aesthetic discriminator such as SVM \cite{nishiyama2009sensation,cheng2010learning}. The release of two cropping databases \cite{yan2013learning,fang2014automatic} further facilitates the training of discriminative cropping models. However, the handcrafted features are not strong enough to accurately predict the complicated image aesthetics \cite{deng2017image}.

\textbf{Data-driven methods.} Most recent methods are data-driven, which train an end-to-end CNN model for image cropping. Limited by the insufficient number of annotated training samples, many methods in this category \cite{chen2017quantitative,wang2017deep,wang2018deep,deng2017image,deng2017aesthetic,guo2017automatic,li2018a2} adopt a general aesthetic classifier trained from image aesthetic databases such as AVA \cite{murray2012ava} and CUHKPQ \cite{luo2011content} to help cropping. However, a general aesthetic classifier trained on full images may not be able to reliably evaluate the crops within one image \cite{chen2017learning,wei2018good}. An alternative strategy is to use pairwise learning to construct more training data \cite{chen2017quantitative,chen2017learning,wei2018good}.

Our method lies in the data-driven category with several advantages over the existing methods. First, we propose a new formulation for image cropping learning. Second, we constructed a much larger scale dataset with reliable annotations, which enables us to train more robust and accurate cropping models.

\subsection{Image cropping efficiency}

Efficiency is important for a practical image cropping system. Two types of efforts can be made to improve the efficiency: reducing the number of candidate crops and decreasing the computational complexity of cropping models. A brutal force sliding window search can easily result in million of candidates for each image. To reduce the number of candidate crops, Wang \textit{et al.} \cite{wang2018deep} first detected the salient region of an image and then generated about 1,000 crops around the salient region. This strategy inevitably suffers from the same problem faced by attention-based methods: many useful background pixels are unnecessarily lost and the cropping results may not have the best composition. Wei \textit{et al.} \cite{wei2018good} employed the pre-defined 895 anchor boxes in the single shot detector (SSD) \cite{liu2016ssd}. Again, the anchor boxes designed for object detection may not be the optimal choice for image cropping. Our new formulation carefully considers the special properties of image cropping and reduces the number of candidate crops to be no more than 90.

Regarding the model complexity, most recent cropping models are based on the AlexNet \cite{chen2017quantitative,chen2017learning} or VGG16 architecture \cite{deng2017image,wang2017deep,wei2018good}, which are too heavy to be deployed on computational resource limited devices such as mobile phones and drones. Our cropping model embraces the latest advances in efficient CNN architecture design and it is much more lightweight and efficient than the previous models.

\section{Grid anchor based image cropping}

As illustrated in Fig. \ref{figure:problems}, image cropping has a high degree of freedom. There is not a unique optimal crop for a given image. We consider two practical requirements of a good image cropping system. Firstly, a reliable cropping system should be able to return acceptable results for different settings (e.g., aspect ratio and resolution) rather than one single output. Secondly, the cropping system should be lightweight and efficient to run on resource limited devices. With these considerations, we propose a grid anchor based formulation for practical image cropping, and construct a new benchmark under this formulation.

\subsection{Formulation}

\begin{figure}[t]
\centering
\subfigure{
\begin{minipage}[b]{1.0\linewidth}
\centering
\includegraphics[width=1.0\textwidth]{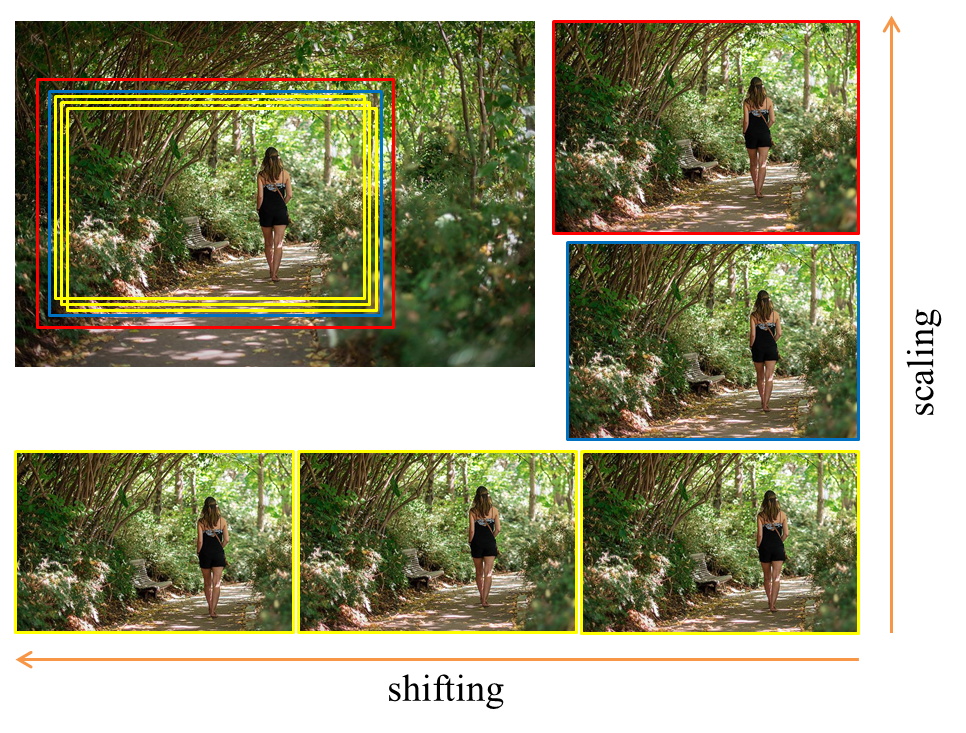}
\end{minipage}}%
\caption{The local redundancy of image cropping. Small local changes (e.g., shifting and/or scaling) on the cropping window of an acceptable crop (the bottom-right one) are very likely to output acceptable crops too.}
\label{figure:localInsensitivity}
\end{figure}

\begin{figure}[t]
\centering
\subfigure{
\begin{minipage}[b]{1.0\linewidth}
\centering
\includegraphics[width=1.0\textwidth]{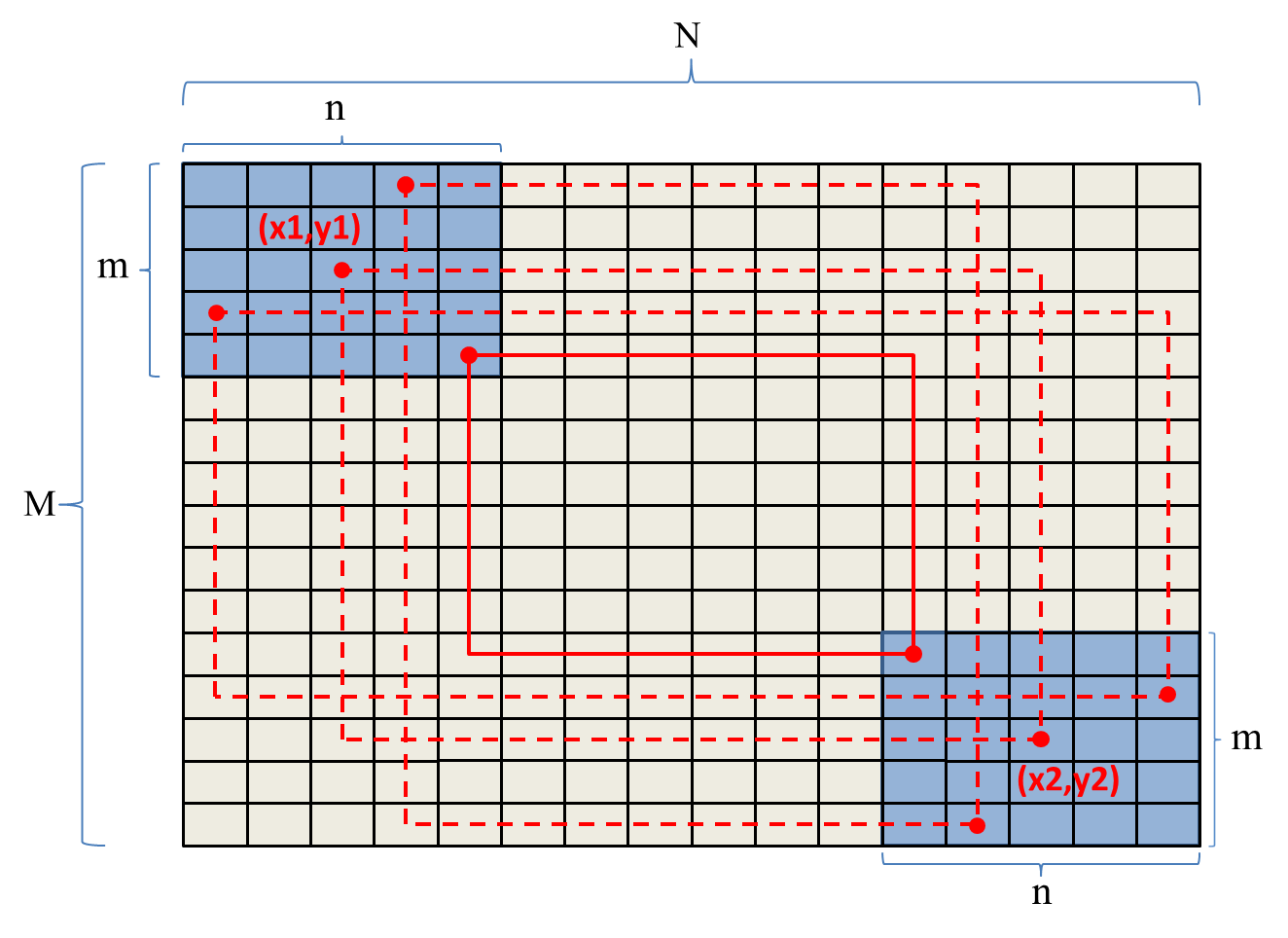}
\end{minipage}}%
\caption{Illustration of the grid anchor based formulation of image cropping. $M$ and $N$ are the numbers of bins for grid partition, while $m$ and $n$ define the adopted range of anchors for content preservation.}
\label{figure:simplification}
\end{figure}

Given an image with spatial resolution $H \times W$, a candidate crop can be defined using its top-left corner $(x_1,y_1)$ and bottom-right corner $(x_2,y_2)$, where $1\leq x_1<x_2\leq H$ and $1\leq y_1<y_2\leq W$. It is easy to calculate that the number of candidate crops is $\frac{H(H-1)W(W-1)}{4}$, which is a huge number even for an image of size $100\times100$. Fortunately, by exploiting the following properties and requirements of image cropping, the searching space can be significantly reduced, making automatic image cropping a tractable problem.

\textit{Local redundancy:} Image cropping is naturally a problem with local redundancy. As illustrated in Fig. \ref{figure:localInsensitivity}, a set of similar and acceptable crops can be obtained in the neighborhood of a good crop by shifting and/or scaling the cropping widow. Intuitively, we can remove the redundant candidate crops by defining crops on image grid anchors rather than dense pixels. The proposed grid anchor based formulation is illustrated in Fig. \ref{figure:simplification}. We construct an image grid with $M \times N$ bins on the original image, and define the corners $(x_1,y_1)$ and $(x_2,y_2)$ of one crop on the grid centers, which serve as the anchors to generate a representative crop in the neighborhood. Such a formulation largely reduces the number of candidate crops from $\frac{H(H-1)W(W-1)}{4}$  to $\frac{M(M-1)N(N-1)}{4}$, which can be several orders smaller.

\begin{figure}[t]
\centering
\subfigure{
\begin{minipage}[b]{1.0\linewidth}
\centering
\includegraphics[width=1.0\textwidth]{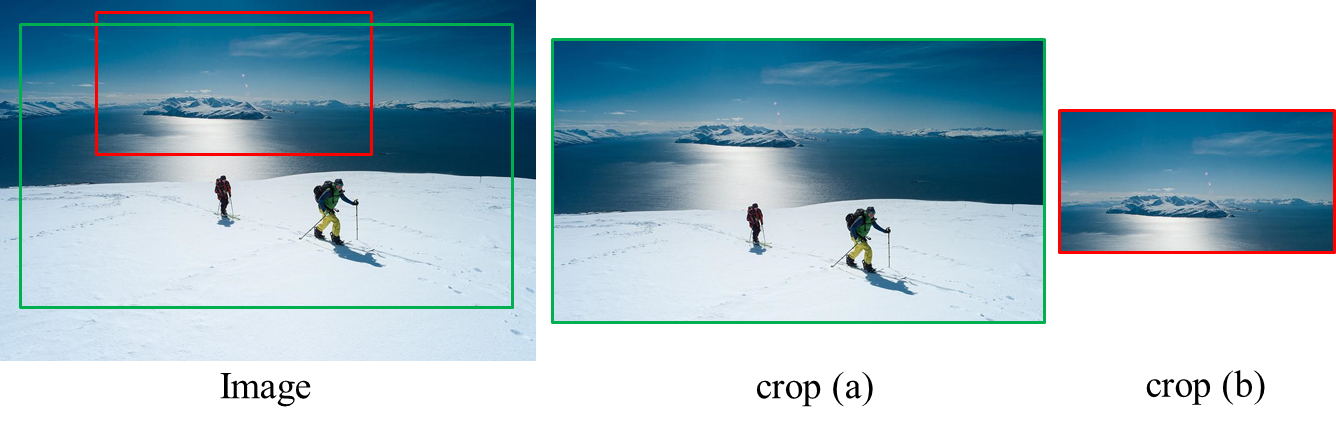}
\end{minipage}}%
\caption{The content preservation of image cropping. The small crop (b) misses the two persons, which are the key objects in the original image although itself has a good composition. With content preservation constraint, crop (a) will be generated to preserve as much useful information as possible.}
\label{figure:contentpreservation}
\end{figure}

\textit{Content preservation:} Generally, a good crop should preserve the major content of the source image \cite{fang2014automatic}. Otherwise, the cropped image may miss important information in the source image and misinterpret the photographer's purpose, resulting in unsatisfied outputs. An example is shown in Fig. \ref{figure:contentpreservation}. As can be seen, without the content preservation constraint, the output crop with good composition may miss the two persons in the scene, which are the key objects in the original image. Therefore, the cropping window should not be too small in order to avoid discarding too much the image content. To this end, we constrain the anchor points $(x_1,y_1)$ and $(x_2,y_2)$ of a crop into two regions with $m \times n$ bins on the top-left and bottom-right corners of the source image, respectively, as illustrated in Fig. \ref{figure:simplification}. This further reduces the number of crops from $\frac{M(M-1)N(N-1)}{4}$ to $m^2n^2$.

The smallest possible crop (highlighted in red solid lines in Fig. \ref{figure:simplification}) generated by the proposed scheme covers about $\frac{(M-2m+1)(N-2n+1)}{MN}$ grids of the source image, which may still be too small to preserve enough image content. We thus further constrain the area of potential crops to be no smaller than a certain proportion of the whole area of source image:
\begin{equation}\label{equ:area constraint}
S_{crop} \geq \lambda S_{Image},
\end{equation}
where $S_{crop}$ and $S_{Image}$ represent the areas of crop and original image, respectively, and $\lambda \in [\frac{(M-2m+1)(N-2n+1)}{MN},1)$.

\textit{Aspect ratio:} Because of the standard resolution of imaging sensors and displays, most people have been accustomed to the popular aspect ratios such as 16:9, 4:3 and 1:1. Candidate crops which have uncommon aspect ratios may be inconvenient to display and can make people feel uncomfortable. We thus require the aspect ratio of acceptable candidate crops satisfy the following condition:
\begin{equation}\label{equ:aspect ratio constraint}
\alpha_1 \leq \frac{W_{crop}}{H_{crop}} \leq \alpha_2,
\end{equation}
where $W_{crop}$ and ${H_{crop}}$ are the width and height of a crop. Parameters $\alpha_1$ and $\alpha_2$ define the range of aspect ratio and we set them to 0.5 and 2 to cover most common aspect ratios.

With Eq. \ref{equ:area constraint} and Eq. \ref{equ:aspect ratio constraint}, the final number of candidate crops in each image is less than $m^2n^2$.

\subsection{Database construction}\label{section: database}

Our proposed grid anchor based formulation reduces the number of candidate crops from $\frac{H(H-1)W(W-1)}{4}$ to less than $m^2n^2$. This enables us to annotate all the candidate crops for each image. To make the annotation cost as low as possible, we first made a small scale subjective study to find the smallest \{$M,N,m,n$\} that ensure at least 3 acceptable crops for each image. We collected 100 natural images and invited five volunteers to participate in this study. We set $M=N$ $\in \{16,14,12,10\}$ and $m=n$ $\in \{5,4,3\}$ to reduce possible combinations. $\lambda$ in Eq. \ref{equ:area constraint} was set to 0.5. After the tests, we found that $M=N=12$ and $m=n=4$ can lead to a good balance between cropping quality and annotation cost. Finally, the number of candidate crops is successfully reduced to no more than 90 for each image. Note that the setting of these parameters mainly aims to reduce annotation cost for training. In the testing stage, it is straightforward to use finer image grid to generate more candidate crops when necessary.

\begin{figure}[t]
\centering
\subfigure{
\begin{minipage}[b]{1.0\linewidth}
\centering
\includegraphics[width=1.0\textwidth]{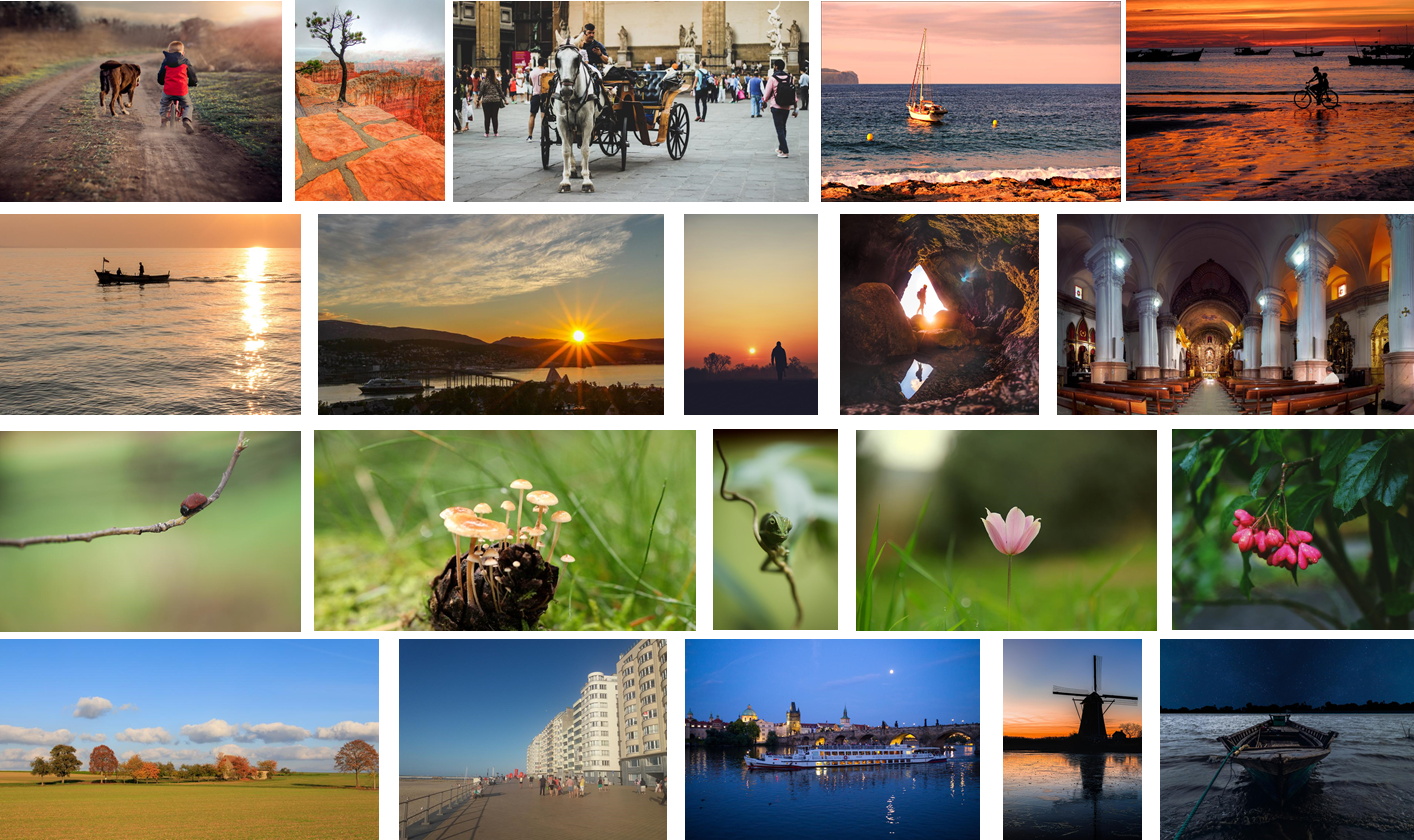}
\end{minipage}}%
\caption{Some sample images from the GAICD dataset.}
\label{figure:database_samples}
\end{figure}

\begin{figure}[t]
\centering
\subfigure{
\begin{minipage}[b]{1.0\linewidth}
\centering
\includegraphics[width=1.0\textwidth]{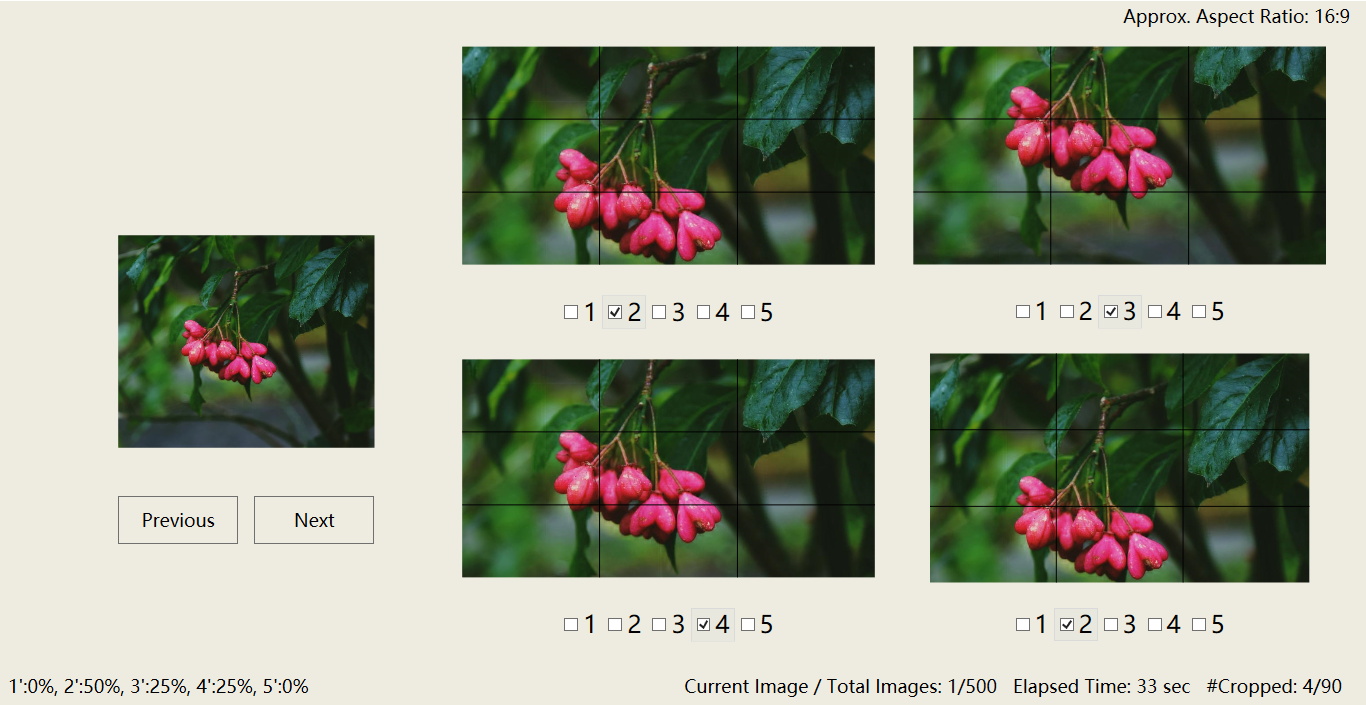}
\end{minipage}}%
\caption{Interface of the developed annotation toolbox.}
\label{figure:annotation}
\end{figure}

\begin{figure}[t]
\centering
\subfigure{
\begin{minipage}[b]{1.0\linewidth}
\centering
\includegraphics[width=1.0\textwidth]{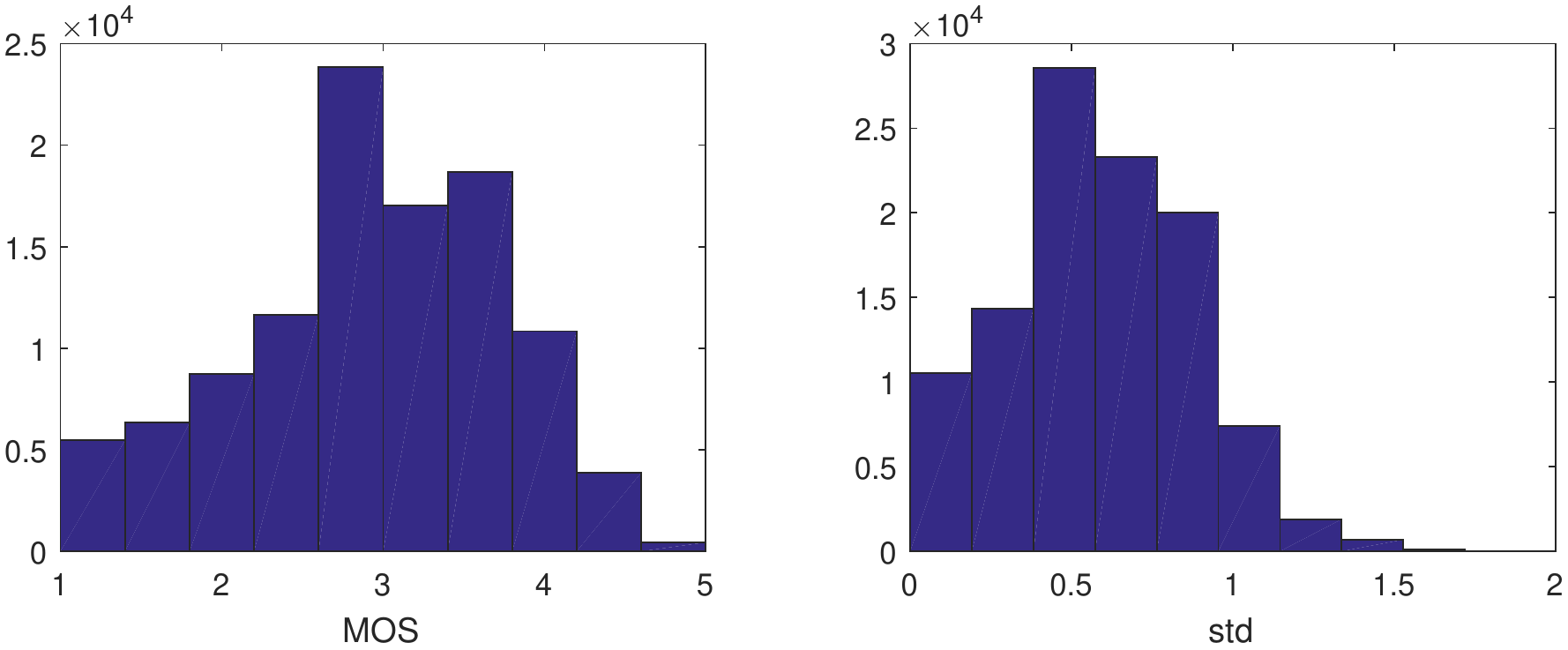}
\end{minipage}}%
\caption{Histograms of the MOS and standard deviation on the GAICD.}
\label{figure:statistics}
\end{figure}

\begin{figure*}[t]
\centering
\subfigure{
\begin{minipage}[b]{1.0\linewidth}
\centering
\includegraphics[width=1.0\textwidth]{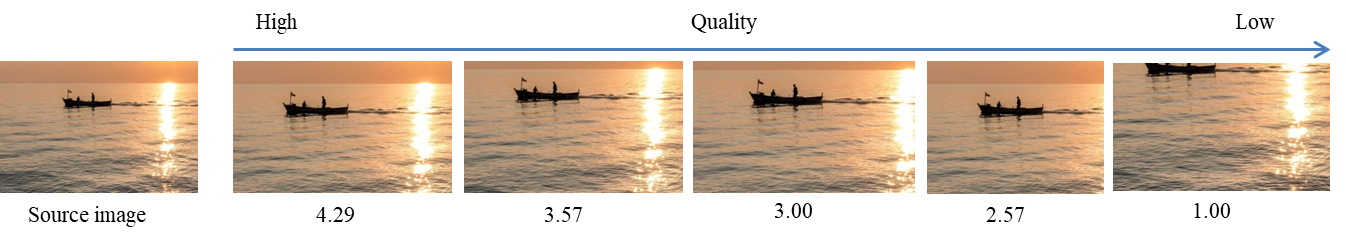}
\end{minipage}}%
\caption{One example source image and several of its annotated crops in our GAICD. The MOS is marked under each crop.}
\label{figure:database_annotation}
\end{figure*}

With the above settings, we constructed a Grid Anchor based Image Cropping Database (GAICD). We first crawled $\sim$50,000 images from the Flickr website. Considering that many images uploaded to Flickr already have good composition, we manually selected 1,000 images whose composition can be obviously improved, as well as 236 images with proper composition to ensure the generality of the GAICD. The selected images have various aspect ratios and cover a variety of scenes and lighting conditions. There are 106,860 candidate crops of the 1,236 images in total. Some sample images from the GAICD are shown in Fig. \ref{figure:database_samples}.

To improve the annotation efficiency, we developed an annotation toolbox whose interface is shown in Fig. \ref{figure:annotation}. Each time, it displays one source image on the left side and 4 crops generated from it on the right side. The crops are displayed in ordered aspect ratio to alleviate the influence of dramatic changes of aspect ratio on human perception. Specifically, we choose six common aspect ratios (including 16:9, 3:2, 4:3, 1:1, 3:4 and 9:16) and group crops into six sets based on their closest aspect ratios. The top-right corner displays the approximate aspect ratio of current crops. Two horizontal and two vertical guidelines can be optionally used to assist judgement during the annotation. For each crop, we provide five scores (from 1 to 5, representing ``bad," ``poor," ``fair," ``good," and ``excellent" crops) to rate by annotators. The annotators can either scroll their mouse or click the ``Previous" or ``Next" buttons to change page. In the bottom-left of the interface, we show the score distribution of rated crops for the current image as a reference for annotators. The bottom-right corner shows the progress of the annotation and the elapsed time.

A total of 19 annotators passed our test on photography composition and participated into the annotation. They are either experienced photographers from photography communities or students from the art department of two universities. Each crop was annotated by seven different subjects. The mean opinion score (MOS) was calculated for each candidate crop as its groundtruth quality score. The histograms of the mean opinion score (MOS) and standard deviation among the 106,860 candidate crops are plotted in Fig. \ref{figure:statistics}. It can be seen that most crops have ordinary or poor quality, while about 10\% crops have MOS larger than 4. Regarding to the standard deviation, only 5.75\% crops are larger than 1, which indicates the consistency of annotations under our grid anchor based formulation. Fig. \ref{figure:database_annotation} shows one source image and several of its annotated crops (with MOS scores) in the GAICD.

Compared to the previous cropping datasets on which only one bounding box or several ranking pairs are annotated, our dataset has much more dense annotation and brings two significant benefits. First, our dense annotation provides not only richer but also finer supervised information for training cropping models. Second, the dense annotation enables us to define more reliable evaluation metrics on our new dataset, providing a more reasonable cropping benchmark for researchers to develop and evaluate their models.

\subsection{Evaluation metrics}\label{Evaluation metrics}

As shown in Table \ref{table:Performance comparison} and Fig. \ref{figure:problems}, the IoU and BDE metrics used in previous studies of image cropping are problematic. The dense annotations of our GAICD enable us to define more reliable metrics to evaluate cropping performance. Specifically, we define three types of metrics on our GAICD. The first tpye of metrics evaluate the ranking correlation between model's predictions and the groundtruth scores; the second type of metrics measure the model's performance to return the best crops; and the third type of metrics consider the ranking information into the best returns.

\textbf{Ranking correlation metrics:} The Pearson correlation coefficient (PCC) \cite{wiki:pcc} and Spearman's rank-order correlation coefficient (SRCC) \cite{wiki:srcc} can be naturally employed to evaluate the model's prediction consistency with the groundtruth MOS in our GAICD. These two metrics have been widely used in image quality and aesthetic assessment \cite{kong2016photo, ma2017waterloo, talebi2018nima}.  Denote by $\mathbf{g}_i$ the vector of MOS of all crops for image $i$, and by $\mathbf{p}_i$ the predicted scores of these crops by a model. The PCC is defined as:
\begin{equation}\label{equ:pcc}
PCC(\mathbf{g}_i,\mathbf{p}_i) = \frac{\textrm{cov}(\mathbf{g}_i,\mathbf{p}_i)}{\sigma_{\mathbf{g}_i}\sigma_{\mathbf{p}_i}},
\end{equation}
where $\textrm{cov}$ and $\sigma$ are the operators of the covariance and standard deviation. One can see that the PCC measures the linear correlation between two variables.

\begin{figure*}[t]
\centering
\subfigure{
\begin{minipage}[b]{1.0\linewidth}
\centering
\includegraphics[width=1.0\textwidth]{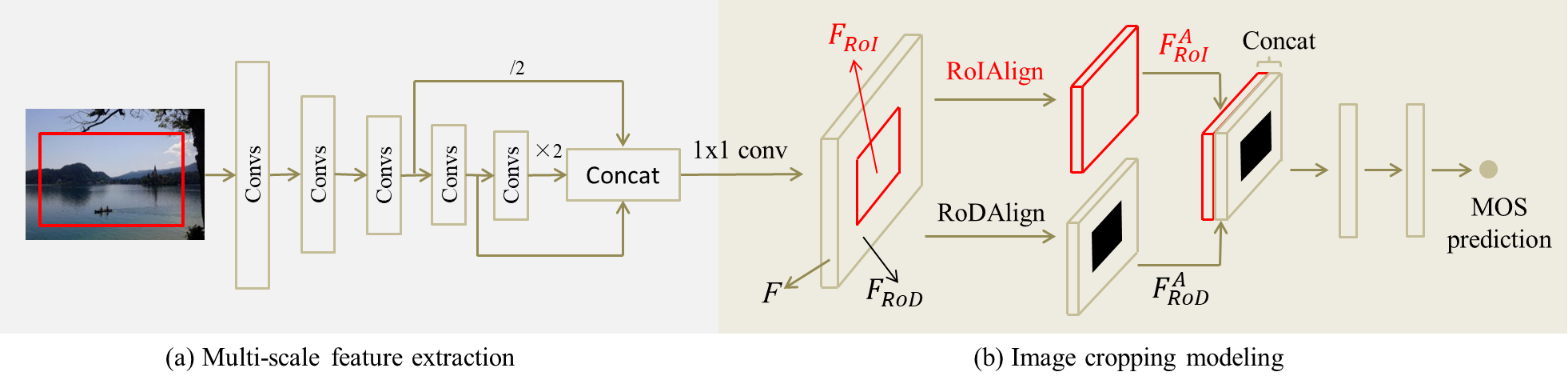}
\end{minipage}}%
\caption{The proposed CNN architecture for image cropping model learning. It consists of a multi-scale feature extraction module and a carefully designed cropping modeling module. Each convolutional block contains several convolution, batch normalization and ReLU layers. Symbols ``$\times 2$" and ``$/2$" represent bilinear upsampling and downsampling, respectively.}
\label{figure:CNN}
\end{figure*}

The SRCC is defined as the PCC between the rank variables:
\begin{equation}\label{equ:srcc}
SRCC(\mathbf{g}_i,\mathbf{p}_i) = \frac{\textrm{cov}(\mathbf{r}_{\mathbf{g}_i},\mathbf{r}_{\mathbf{p}_i})}{ \sigma_{\mathbf{r}_{\mathbf{g}_i}}\sigma_{\mathbf{r}_{\mathbf{p}_i}}},
\end{equation}
where $\mathbf{r}_{\mathbf{g}_i}$ and $\mathbf{r}_{\mathbf{p}_i}$ record the ranking order of scores in $\mathbf{g}_i$ and $\mathbf{p}_i$, respectively. The SRCC assesses the monotonic relationship between two variables. Given a testing set with $T$ images, we calculate the average PCC and average SRCC over the $T$ images as the final results:
\begin{equation}\label{equ:avg srcc}
\overline{PCC} = \frac{1}{T}\sum\nolimits_{i=1}^{T} PCC(\mathbf{g}_i,\mathbf{p}_i),
\end{equation}
\begin{equation}\label{equ:avg srcc}
\overline{SRCC} = \frac{1}{T}\sum\nolimits_{i=1}^{T} SRCC(\mathbf{g}_i,\mathbf{p}_i).
\end{equation}

\textbf{Best return metrics:} Considering that in practical cropping applications, users care more about whether the cropping model can return the best crops rather than accurately rank all the candidate crops, we define a set of metrics to evaluate the models' ability to return the best crops. This new set of metrics is called as ``return $K$ of top-$N$" accuracy, which is similar to the ``Precision at K" metric \cite{wiki:precisionK} widely used in modern retrieval systems. Specifically, we define the best crops of image $i$ as the set of crops whose MOS rank top-$N$, and we denote this top-$N$ set by $S_{i}(N)$. Suppose a cropping model returns $K$ crops that have the highest prediction scores. We denote these $K$ crops by $\{c_{ik}\}_{k=1}^K$ for image $i$. The ``return $K$ of top-$N$" accuracy checks on average how many of the returned $K$ crops fall into the top-$N$ set $S_{i}(N)$. It is defined as:
\begin{equation}\label{equ:topN accuracy}
Acc_{K/N} = \frac{1}{TK}\sum\nolimits_{i=1}^{T}\sum\nolimits_{k=1}^{K} True(c_{ik} \in S_{i}(N)),
\end{equation}
where $True(*) = 1$ if * is true, otherwise $True(*) = 0$. In practice, the number $K$ of returned crops should not be set too large for users' convenience. In the ideal case, a cropping model should return only 1 crop to meet the user's expectation. For more patient users, no more than 4 crops could be returned to them. We thus set $K$ to 1, 2, 3 and 4 in our benchmark. Regarding the selection of $N$, the statistic of MOS discussed in Section \ref{section: database} shows that about 10\% crops have MOS larger than 4, which means that there are on average 10 good crops for each image. We thus set $N$ to 5 or 10. As a result, we obtain 8 accuracy indexes $Acc_{K/N}$ based on the different combinations of $K$ and $N$.

\textbf{Rank weighted best return metrics:} The metric $Acc_{K/N}$ does not distinguish the rank among the returned top-$N$ crops. For example, the value of $Acc_{1/5}$ will be the same when returning either the rank-1 or rank-5 crop. To further distinguish the rank of the returned top-$N$ crops, we introduce a set of rank weighted best return metrics. Given the returned $K$ crops of image $i$ and their ranks among all the candidate crops, denoted by $\{r_{ik}\}_{k=1}^K$, we sort the $K$ crops to have descending MOS, and obtain the sorted $K$ crops $\{c_{ij}\}_{j=1}^K$ associated with their ranks $\{r_{ij}\}_{j=1}^K$. The ``rank weighted return $K$ of top-$N$" accuracy is defined as:
\begin{equation}\label{equ:weighted topN accuracy}
Acc^w_{K/N} = \frac{1}{TK}\sum\nolimits_{i=1}^{T}\sum\nolimits_{j=1}^{K} True(c_{ij} \in S_{i}(N))\ast w_{ij},
\end{equation}
where
\begin{equation}\label{equ:weight in accuracy}
w_{ij} = e^{\frac{-\beta(r_{ij}-j)}{N}},
\end{equation}
where $\beta>0$ is a scaling parameter and we simply set it to 1.
The weight $w_{ij}$ is designed under two considerations. First, $w_{ij}$ should be larger if the crop $c_{ij}$ has better rank. Second, $w_{ij}$ should be 1 if the sorted rank $r_{ij}$ matches the order of $c_{ij}$ among the $K$ returns, making the rank weighted accuracy $Acc^w_{K/N}$ able to reach 1 when the best crop set is returned.

We give an example to illustrate the calculation of $Acc^w_{4/5}$ for an input image. Suppose the returned 4 crops are ranked as $\{r_{ik}\}_{k=1}^K=\{2,5,3,10\}$ among all candidate crops, it is easy to have $Acc_{4/5}$ = 0.75 since three are 3 crops falling into the top-$5$ set. The sorted ranks of the four returns are $\{r_{ij}\}_{j=1}^K=\{2,3,5,10\}$, and the rank weighted accuracy is calculated as $Acc^w_{4/5}=\frac{1}{4}(e^{-\frac{2-1}{5}}+e^{-\frac{3-2}{5}}+e^{-\frac{5-3}{5}})=0.5769$. Compared with $Acc_{K/N}$, the metric $Acc^w_{K/N}$ can more precisely distinguish the quality of returns.

\section{Cropping model learning}

Limited by the insufficient amount of training data, most previous cropping methods focused on how to leverage additional aesthetic databases \cite{wang2017deep,chen2017learning,deng2017aesthetic} or how to construct more training pairs \cite{chen2017quantitative,wei2018good}, paying limited attention to how to design a more suitable network for image cropping itself. They usually adopt the standard CNN architecture widely used in object detection. Our GAICD provides a better platform with much more annotated samples for model training. By considering the special properties of image cropping, we design an effective and lightweight cropping model. The overall architecture is shown in Fig. \ref{figure:CNN}, which consists of a multi-scale feature extraction module and a carefully designed image cropping module. We also employ a set of data augmentation operations for learning robust cropping models.

\subsection{Multi-scale feature extraction module}

\textbf{Efficient base model:} A practical cropping model needs to be lightweight and efficient enough to be deployed on resource limited devices. Instead of employing those classical pre-trained CNN models such as AlexNet \cite{krizhevsky2012imagenet}, VGG16 \cite{simonyan2014very} or ResNet50 \cite{he2016deep} as in previous work \cite{wang2017deep,deng2017image,chen2017quantitative,chen2017learning,deng2017aesthetic,guo2017automatic,li2018a2,wei2018good,zeng2019reliable}, we choose the more efficient architectures including the MobileNetV2 \cite{sandler2018MobileNetV2} and ShuffleNetV2 \cite{ma2018shufflenet}. Fortunately, we found that using such efficient models will not sacrifice the cropping accuracy compared with their complicated counterparts, mostly owing to the special properties of image cropping and more advanced architecture designs of the MobileNetV2 and ShuffleNetV2. More details and discussions can be found in the ablation experiments.

\textbf{Multi-scale features:} As illustrated by the two examples in Fig. \ref{figure:ruleofthirds}, the scale of objects varies significantly in different scenes. The features should also be responsive to the local distracting contents which should be removed in the final crop. As shown in Fig. \ref{figure:CNN}(a), we extract multi-scale features from the same backbone CNN model. It has been widely acknowledged that the shallower CNN layers tend to capture the local textures while the deeper layers model the entire scene \cite{zeiler2014visualizing}. This motivates us to concatenate the feature maps from three different layers. Since the feature maps in different layers have different spatial resolution, we use bilinear downsampling and upsampling to make them have the same spatial resolution.
The three feature maps are concatenated along the channel dimension as the output feature map.

\subsection{Cropping module}


\textbf{Modeling both the RoI and RoD:} One special property of image cropping is that we need to consider not only the region of interest (RoI) but also the region to be discarded (hereafter we call it region of discard (RoD)). On one hand, removing distracting information can significantly improve the composition. On the other hand, cutting out important region can dramatically change or even destroy an image. Taking the second last crop in Fig. \ref{figure:database_annotation} as an example, although it may have acceptable composition, its visual quality is much lower than the source image because the beautiful sunset glow is cropped out. The discarded information is unavailable to the cropping model if only the RoI is considered, while modeling the RoD can effectively solve this problem.

Referring to Fig. \ref{figure:CNN}, denote by $F$ the whole feature map output by the feature extraction module, and denote by $F_{RoI}$ and $F_{RoD}$ the feature maps in RoI and RoD, respectively. We first employ the RoIAlign operation \cite{he2017mask} to transform $F_{RoI}$ into $F_{RoI}^{A}$, which has fixed spatial resolution $s \times s$. The $F_{RoD}$ is constructed by removing $F_{RoI}$ from $F$, namely, setting the values of $F_{RoI}$ to zeros in $F$. Then the RoDAlign operation (using the same bilinear interpolation as RoIAlign) is performed on $F_{RoD}$, resulting in $F_{RoD}^{A}$ which has the same spatial resolution as $F_{RoI}^{A}$. $F_{RoI}^{A}$ and $F_{RoD}^{A}$ are concatenated along the channel dimension as one aligned feature map which contains the information of both RoI and RoD. The combined feature map is fed into two fully connected layers for final MOS prediction. Throughout our experiments, we fix $s$ as 9 so that the bilinear interpolation in RoIAlign and RoDAlign can effectively leverage the entire feature map output by our feature extraction module. To be more specific, an input image of resolution $256 \times 256$ results in $16 \times 16$ feature maps after our feature extraction module, and bilinear interpolation takes four points to interpolate one point.

\begin{figure}[t]
\centering
\subfigure{
\begin{minipage}[b]{1.0\linewidth}
\centering
\includegraphics[width=1.0\textwidth]{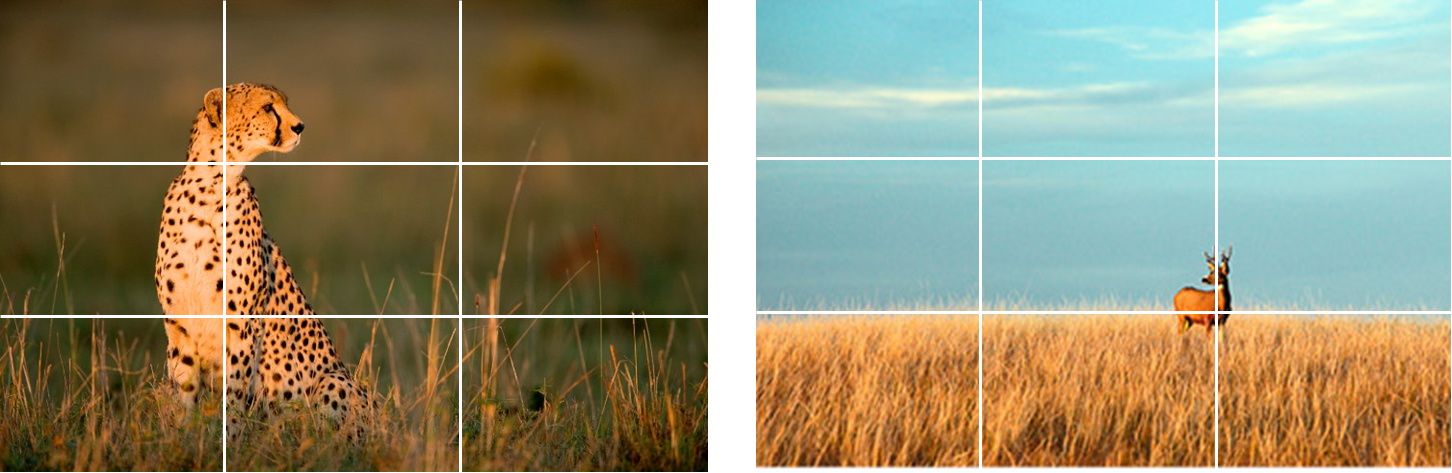}
\end{minipage}}%
\caption{Two examples using ``rule of thirds" \cite{wiki:rule-of-thirds} composition.} 
\label{figure:ruleofthirds}
\end{figure}

\textbf{Modeling the spatial arrangement:} The spatial arrangement of context and objects in an image plays a key role in image composition. For example, the most commonly used ``rule of thirds" composition rule suggests to place important compositional elements at certain locations of an image \cite{wiki:rule-of-thirds}. Specifically, an image can be divided into nine parts by two equally spaced horizontal lines and two equally spaced vertical lines, and important elements should be placed along these lines or at the intersections of these lines, as shown in Fig \ref{figure:ruleofthirds}. Other common composition rules such as symmetry and leading line also have certain spatial pattern. Considering that the downsampling and pooling operations after the feature extraction stage can cause significant loss of spatial information, we employ a fully-connected layer with large kernel size to explicitly model the spatial arrangement of an image. Our experimental results validate the advantage of this design in both cropping accuracy and efficiency than using several convolutional layers.

\textbf{Reducing the channel dimension:} Another characteristic of image cropping is that it does not need to accurately recognize the category of different objects or scenes, which allows us to significantly reduce the channel dimension of the feature map. In practice, we found that the feature channel dimension can be reduced from several hundred to only 8 by using $1 \times 1$ convolution without sacrificing much the performance. The low channel dimension makes our image cropping module very efficient and lightweight.

\textbf{Loss function:} Denote by $e_{ij}=g_{ij}-p_{ij}$, where $g_{ij}$ and $p_{ij}$ are the groundtruth MOS and predicted score of the $j$-th crop for image $i$. The Huber loss \cite{huber1964robust} is employed as the loss function to learn our cropping model because of its robustness to outliers:
\begin{equation}\small \label{equ:Huber}
\mathcal{L}_{ij}=
\left\{
\begin{aligned}
&\frac{1}{2}e_{ij}^2, \textrm{when}\; |e_{ij}|\leq\delta,\\
&\delta|e_{ij}|-\frac{1}{2}\delta^2, \textrm{otherwise},
\end{aligned}
\right.
\end{equation}
where $\delta$ is fixed at 1 throughout our experiments.

\subsection{Data augmentation}

Data augmentation is an effective way to improve the robustness and performance of deep CNN models. However, many popular data augmentation operations are inappropriate for cropping. For example, rotation and vertical flipping can severely destroy the composition. Since the IoU is unreliable for evaluating cropping performance, randomly generating crops and assigning labels to them based on IoU \cite{chen2017learning} is also questionable. We thus employ a set of operations, which do not affect the composition, for data augmentation. Specifically, we randomly adjust the brightness, contrast, saturation, hue and horizontally flip the input image in the training stage.

\section{Experiments}

\begin{table*}[t]
\centering
\caption{Image cropping performance by using different feature extraction modules. The FLOPs are calculated on image with $256\times 256$ pixels. All the single-scale models use the feature map after the fourth convolutional block which has the best performance among the three scales.  The last convolutional block in both the MobileNetV2 and ShuffleNetV2 contains most of the parameters because of the high channel dimension, while it is simply a max pooling layer in VGG16 model and does not have any parameter.}
\label{table:model}
\begin{tabular}{|c|cccc|c|c|c|c|c|c|c|c|c|c|}
\hline
& & & & & & & & & & & &\\[-1em]
Base model  & Scale & Aug. & FLOPs & \# of params & $\overline{SRCC}$ & $\overline{PCC}$ & $Acc_{1/5}$ & $Acc_{4/5}$ & $Acc_{1/10}$ & $Acc_{4/10}$ & $Acc^w_{4/5}$  & $Acc^w_{4/10}$  \\\hline
\multirow{3}{*}{VGG16}
&Single&No &22.3G&14.7M&0.752&0.778&58.0&47.7&74.0&67.9&32.2&49.2  \\
&Single&Yes&22.3G&14.7M&0.764&0.791&59.5&49.2&76.0&69.3&33.3&50.3  \\
&Multi &Yes&22.3G&14.7M& \textbf{0.777} &\textbf{0.800}&\textbf{60.5}&\textbf{50.2} &\textbf{77.5}& \textbf{70.6}&\textbf{34.4}&\textbf{51.3} \\\hline
\multirow{3}{*}{MobileNetV2}
&Single&No &314M &0.54M&0.760&0.782&58.5&49.1&75.5&69.0&33.6&51.6  \\
&Single&Yes&314M &0.54M&0.775&0.793&60.5&51.4&77.5&70.9&35.1&53.2  \\
&Multi &Yes&407M &1.81M& \textbf{0.783} &\textbf{0.806}&\textbf{62.5}&\textbf{52.5} &\textbf{78.5}& \textbf{72.3} &\textbf{36.2} &\textbf{54.4}\\\hline
\multirow{3}{*}{ShuffleNetV2}
&Single&No &126M&0.28M&0.751&0.780&58.0&48.8&76.0&68.1&34.2&51.1\\
&Single&Yes&126M&0.28M&0.763&0.792&60.0&50.9&77.5&70.3&35.8&52.6  \\
&Multi &Yes&170M&0.78M&\textbf{0.774} &\textbf{0.801}&\textbf{61.5}& \textbf{52.0} &\textbf{78.5}& \textbf{71.3}&\textbf{37.2}&\textbf{53.6} \\\hline
\end{tabular}
\end{table*}

\begin{figure*}[t]
\centering
\subfigure{
\begin{minipage}[b]{1.0\linewidth}
\centering
\includegraphics[width=1.0\textwidth]{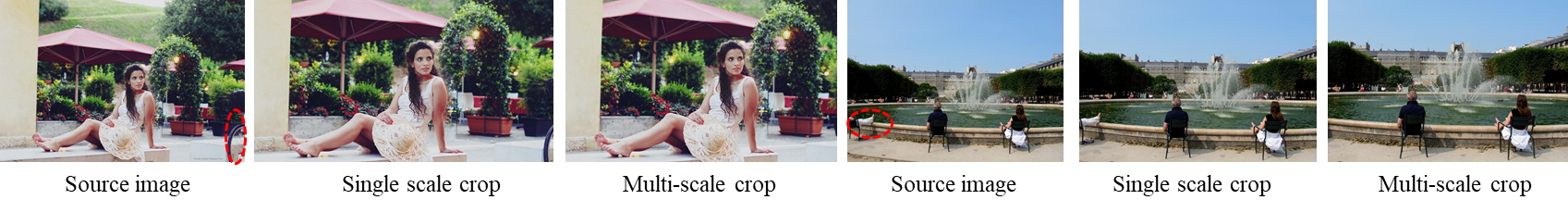}
\end{minipage}}%
\caption{Qualitative comparison between single-scale and multi-scale feature based crops. Using multi-scale features can effectively detect and remove local distracting elements that tend to be ignored by single-scale feature.}
\label{figure:multiscale}
\end{figure*}

\begin{figure*}[t]
\centering
\subfigure{
\begin{minipage}[b]{1.0\linewidth}
\centering
\includegraphics[width=1.0\textwidth]{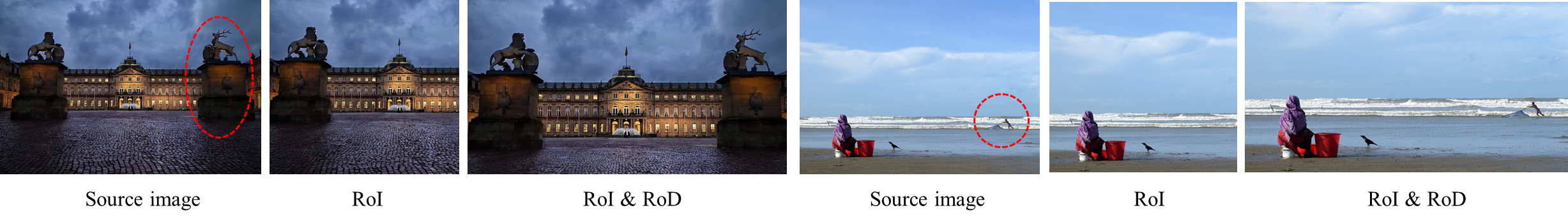}
\end{minipage}}%
\caption{Qualitative comparison of modeling only RoI against modeling both RoI and RoD. Modeling both RoI and RoD can preserve as much useful information as possible in the source image, while modeling only the RoI may lead to unnecessary information loss.}
\label{figure:roirod}
\end{figure*}

\subsection{Implementation details}

We randomly selected 200 images from our GAICD as the testing set and used the remaining 1,036 images (containing 89,519 annotated crops in total) for training and validation. In the training stage, our model takes one image and 64 randomly selected crops of it as a batch to input. In the testing stage, the trained model evaluates all the generated crops of one image and outputs a predicted MOS for each crop. To improve the training and testing efficiency, the short side of input images is resized to 256. The feature extraction module employs the CNN models pre-trained on the ImageNet dataset. The cropping modeling module is randomly initialized using the method proposed in \cite{glorot2010understanding}. The standard ADAM \cite{kingma2014adam} optimizer with the default parameters is employed to train our model for 80 epoches. Learning rate is fixed at $1e^{-4}$ throughout our experiments. The RGB values in input images are scaled to the range of [0,1] and normalized using the mean and standard deviation calculated on the ImageNet. The MOS are normalized by removing the mean and dividing by the standard deviation across the training set. More implementation details can be found in our released code.

\subsection{Ablation study}

\subsubsection{Feature extraction module}\label{Feature extraction module}

We first conduct a set of ablation studies to evaluate the performance of different base models for feature extraction, single-scale and multi-scale features and data augmentation for model training.  The three different base models include VGG16 \cite{simonyan2014very}, MobileNetV2 \cite{sandler2018MobileNetV2} and ShuffleNetV2 \cite{ma2018shufflenet}. The width multiplier is set to 1.0 for both MobileNetV2 and ShuffleNetV2. In these experiments, the image cropping module (including both the RoI and RoD) is fixed for all cases except that the $1 \times 1$ convolutional layer for dimension reduction has different input dimension for different models. Since the accuracy indexes have similar tendency, we only report several representative indexes for each type of metrics in the ablation study to save space. The specific setting, model complexity and cropping performance for each case are reported in Table \ref{table:model}.

\textbf{Base model:} We found that the lightweight MobileNetV2 and ShuffleNetV2 obtain even better performance than the VGG16 model on all the three types of evaluation metrics. This is owning to the more advanced architecture designs of MobileNetV2 and ShuffleNetV2, both of which leverage many latest useful practices in CNN architecture design such as residual learning, batch normalization and group convolution. It is worth mentioning that their classification accuracies are also comparable or slightly better than the VGG16 model on the ImageNet dataset. Besides, the lightweight networks of MobileNetV2 and ShuffleNetV2 are easier to be trained than VGG16 considering the fact that our GAICD is still not very big. Between MobileNetV2 and ShuffleNetV2, the former obtains slightly better performance (at the same width multiplier) on most of the metrics, which is consistent to their relative performances in other vision tasks \cite{ma2018shufflenet}. Regarding the computational cost, both the number of parameters and FLOPs (the number of multiply-adds) of MobileNetV2 and ShuffleNetV2 are more than one order smaller the the VGG16.

\textbf{Multi-scale features:} As can been seen from Table \ref{table:model}, extracting multi-scale features improves the performance for all the three base models. As we have mentioned before, we extract three scales of features from the same backbone network. The single-scale models employed in this study only used the feature map after the fourth convolutional block which was found to have the best performance among the three scales. Regarding the computational cost, extracting multi-scale features only needs to calculate one additional (i.e. the fifth) convolutional block, and the FLOPs only increase by about $\frac{1}{3}$ compared with the single-scale counterparts. Thus the multi-scale models are still very lightweight and efficient. A qualitative comparison of the cropping results by single- and multi-scale features on two images is shown in Fig. \ref{figure:multiscale}. The cropping results show that using multi-scales features can effectively remove local distracting elements that may be ignored by single-scale features.

\begin{figure}[t]
\centering
\subfigure{
\begin{minipage}[b]{1.0\linewidth}
\centering
\includegraphics[width=0.9\textwidth]{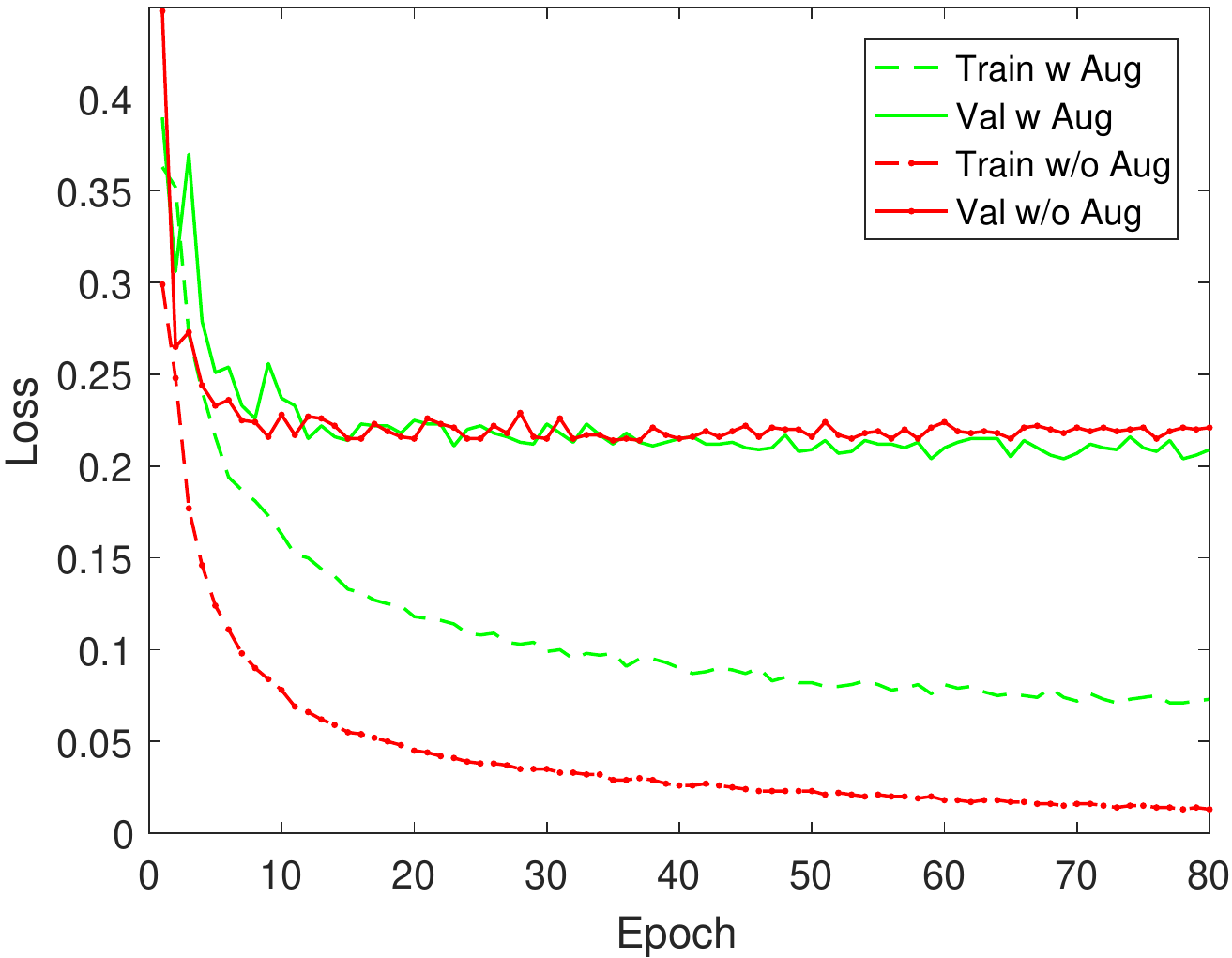}
\end{minipage}}%
\caption{Learning curves with and without data augmentation by the ShuffleNetV2.}
\label{figure:augmentation}
\end{figure}

\textbf{Data augmentation:} The results in Table \ref{table:model} show that data augmentation consistently improves the performance in terms of all the employed metrics for all the three base models. The learning curves with and without data augmentation by ShuffleNetV2 are plotted in Fig. \ref{figure:augmentation}, where we randomly selected 100 images from the training set for validation. One can see that, without data augmentation, the loss decreases very fast on the training set but it has a significant gap to the loss on the validation set. By using data augmentation, smaller loss can be obtained on the validation set, improving the generalization capability of the trained model.

\subsubsection{Image cropping module}

\begin{table*}[t]
\footnotesize
\centering
\caption{Ablation experiments on the roles of RoI and RoD.}
\label{table:roirod}
\begin{tabular}{|c|c|c|c|c|c|c|c|c|c|c|c|}
\hline
 & & & & & & & & & & & \\[-1em]
Base model&module& $\overline{SRCC}$&$\overline{PCC}$& $Acc_{1/5}$ & $Acc_{4/5}$ & $Acc_{1/10}$ & $Acc_{4/10}$ & $Acc^w_{1/5}$ & $Acc^w_{4/5}$ & $Acc^w_{1/10}$& $Acc^w_{4/10}$  \\\hline
\multirow{3}{*}{MobileNetV2}
&RoD         & 0.672 & 0.715 & 45.0 & 39.8 & 61.0 & 56.6 &31.9 &26.4 & 43.2 & 41.3  \\
&RoI         & 0.770 & 0.792 & 60.5 & 51.4 & 76.5 & 71.1 &37.1 &34.6 & 55.3 & 52.4\\
&RoI+RoD     & \textbf{0.783} &\textbf{0.806}&\textbf{62.5}&\textbf{52.5} &\textbf{78.5}& \textbf{72.3} &\textbf{39.6}&\textbf{36.2} &\textbf{56.9}&\textbf{54.4}   \\\hline
\multirow{3}{*}{ShuffleNetV2}
&RoD         & 0.678 & 0.718 & 45.0 & 39.1&61.5&55.7&32.4&28.0&44.6&41.7  \\
&RoI         & 0.764 & 0.785 & 59.5 & 50.1&76.5&69.6&39.2&35.4&55.4&51.6  \\
&RoI+RoD     &\textbf{0.774} &\textbf{0.801}&\textbf{61.5}& \textbf{52.0} &\textbf{78.5}& \textbf{71.3}&\textbf{40.3} &\textbf{37.2}&\textbf{57.3}&\textbf{53.6}   \\\hline
\end{tabular}
\end{table*}

We then evaluate the three special designs in the proposed image cropping module, including RoI and RoD modeling, large kernel size and low channel dimension.

\textbf{RoI and RoD:} We evaluate the roles of RoI and RoD on both MobileNetV2 and ShuffleNetV2 with all the other settings fixed. The results of modeling only RoI, only RoD and both of them are reported in Table \ref{table:roirod}. As can be seen, modeling only the RoD obtains unsatisfied performance, modeling only the RoI performs much better, while modeling simultaneously the RoI and RoD achieves the best cropping accuracy in all cases. A qualitative comparison of modeling only RoI against modeling both RoI and RoD is shown in Fig. \ref{figure:roirod}. One can observe that modeling both RoI and RoD can preserve as much useful information as possible in the source image while modeling only the RoI may lead to some information loss. This corroborates our analysis that image cropping needs to consider both the RoI and RoD.

\textbf{Kernel size:} Given the feature map after RoIAlign and RoDAlign with $9\times 9$ spatial resolution, we propose to use one fully-connected (FC) layer with a large kernel to explicitly modeling the spatial arrangement of the feature map rather than using several small size stride convolutional (Conv) layers. A comparison of using one single $9\times 9\times 16\times768$ FC layer, two $5\times 5\times 16\times768$ Conv layers (the first layer uses stide 2, followed by a $1\times1\times768\times16$ Conv layer for dimension reduction) and three $3\times 3\times 16\times768$ Conv layers (the first two layers use stide 2, each followed by a $1\times1\times768\times16$ Conv layer) are listed in the top half of Table \ref{table:cropping module}. One can see that using one single FC layer obtains higher performance than its competitors. This is because the spatial information may be lost in the downsampling process by stride convolution. Regarding the computational cost, using one single FC layer also has smaller FLOPs since each element calculates only once, while the feature map is repeatedly calculated in the other two cases.

\textbf{Channel dimension reduction:} We also evaluate a set of channel dimensions for the FC layer and report the results in the bottom half of Table \ref{table:cropping module}. Given the multi-scale feature map output by ShuffleNetV2 with 812 channels, we can reduce the channel dimension to only 8 with little performance decay. Note that the channel dimension in the kernel is double of the that in the feature map because of concatenation of the RoI and RoD branches. The performance is still reasonable even if we reduce the channel dimension to 1. Benefiting from the low channel dimension, the FLOPs of our cropping modeling module is only 1.0M, which is almost ignorable compared to the FLOPs in feature extraction.

\begin{table*}[t]
\footnotesize
\centering
\caption{Image cropping performance by using different number and size of kernels in the cropping modeling module. The ShuffleNetV2 model is employed as the feature extraction module for all cases. Note that the channel dimension in the kernel is double of that in the feature map because of concatenation of the RoI and RoD branches.}
\label{table:cropping module}
\begin{tabular}{|c|c|c|c|c|c|c|c|c|c|c|c|}
\hline
 & & & & & & & & & & & \\[-1em]
kernels & FLOPs & $\overline{SRCC}$& $\overline{PCC}$ & $Acc_{1/5}$ & $Acc_{4/5}$ & $Acc_{1/10}$ & $Acc_{4/10}$ & $Acc^w_{1/5}$ & $Acc^w_{4/5}$ & $Acc^w_{1/10}$& $Acc^w_{4/10}$  \\\hline
 & & & & & & & & & & & \\[-1em]
$\left[\begin{array}{c}3,3,16,768 \end{array}\right]\times 3$   & 4.28M  & 0.765 &0.785&57.5&48.6&74.0&68.8&37.1&33.4&53.0&49.8     \\[0.2em]
$\left[\begin{array}{c}5,5,16,768 \end{array}\right]\times 2$   &  8.28M  &0.769 &0.795&60.5&51.2&76.5&70.1&38.7&35.1&55.9&52.4     \\[0.2em]
$\left[\begin{array}{c}9,9,16,768 \end{array}\right]\times 1$   &  1.00M  &\textbf{0.774} &\textbf{0.801}&\textbf{61.5}& \textbf{52.0} &\textbf{78.5}& \textbf{71.3}&\textbf{40.3} &\textbf{37.2}&\textbf{57.3}&\textbf{53.6}  \\[0.2em]\hline
 & & & & & & & & & & & \\[-1em]
$\left[\begin{array}{c}9,9,64,768 \end{array}\right]\times 1$  & 3.98M  & \textbf{0.780} & \textbf{0.806} & \textbf{62.5} & \textbf{52.5} & 79.0 & \textbf{71.8} & 40.5 & 37.4 & \textbf{57.7} & \textbf{54.1}\\[0.2em]
$\left[\begin{array}{c}9,9,32,768 \end{array}\right]\times 1$  & 1.99M  &0.777 &0.804&62.0& 52.2 &\textbf{79.5}&71.5 &\textbf{40.7}&\textbf{37.5}&57.5&53.8\\[0.2em]
$\left[\begin{array}{c}9,9,16,768 \end{array}\right]\times 1$  & 1.00M  &0.774 &0.801&61.5& 52.0 &78.5& 71.3&40.3&37.2&57.3&53.6    \\[0.2em]
$\left[\begin{array}{c}9,9,8,768 \end{array}\right]\times 1$   & 0.50M  & 0.767&0.793&62.0& 51.6 &78.0& 70.7&39.5&36.6&56.8&53.1 \\[0.2em]
$\left[\begin{array}{c}9,9,4,768 \end{array}\right]\times 1$   & 0.25M  & 0.760&0.785&61.0& 50.7 &77.0& 69.5&38.6&35.5&56.1&52.1  \\[0.2em]
$\left[\begin{array}{c}9,9,2,768 \end{array}\right]\times 1$   & 0.13M  & 0.752&0.775&59.0& 48.4 &75.0& 67.5&37.1&34.1&54.8&50.3    \\[0.2em]\hline
\end{tabular}
\end{table*}

\subsection{Comparison to other methods}


\begin{figure*}[t]
\centering
\subfigure{
\begin{minipage}[b]{1.0\linewidth}
\centering
\includegraphics[width=1.0\textwidth]{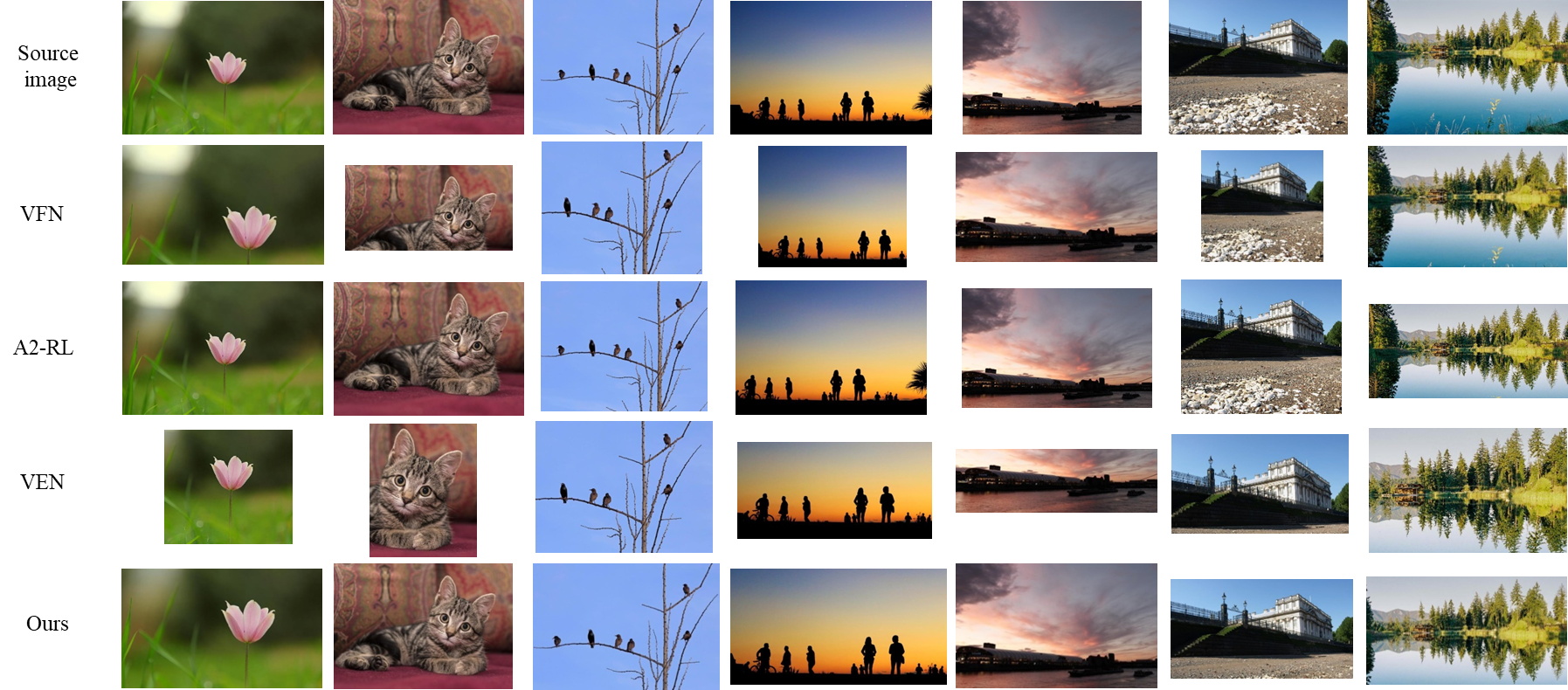}
\end{minipage}}%
\caption{Qualitative comparison of returned top-1 crops by different methods.}
\label{figure:qua comp top1}
\end{figure*}

\begin{figure*}[t]
\centering
\subfigure{
\begin{minipage}[b]{1.0\linewidth}
\centering
\includegraphics[width=1.0\textwidth]{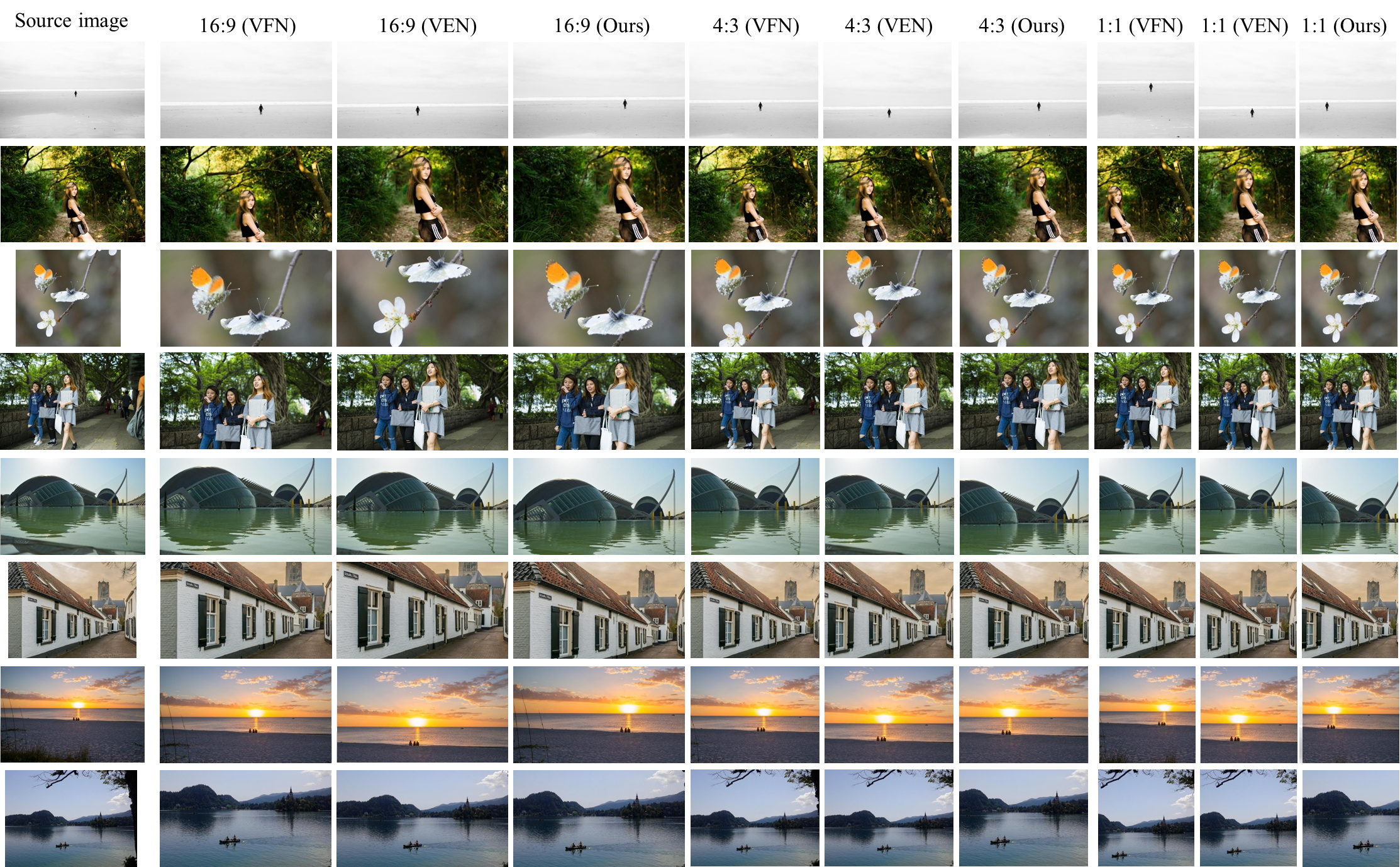}
\end{minipage}}%
\caption{Qualitative comparison of returning crops with different aspect ratios by different methods.}
\label{figure:qua comp fixed}
\end{figure*}

\begin{table*}[t]
\footnotesize
\centering
\caption{Quantitative comparison between different methods on the GAICD. ``--" means that the result is not available. The reported FPS are tested on our own devices using one GTX 1080Ti GPU and i7-6800K CPU. [$\star$] The GPU speeds are inconsistent with the CPU speeds because some operations such as group convolution and channel shuffle in the MobileNetV2 and ShuffleNetV2 are not well supported in PyTorch to make full use of the GPU computational capability.}
\label{table:others}
\begin{tabular}{|c|cccc|cccc|cc|}
\hline
& & & & & & & & & & \\[-1em]
Method                     & $Acc_{1/5}$ & $Acc_{2/5}$ & $Acc_{3/5}$ & $Acc_{4/5}$ & $Acc_{1/10}$ & $Acc_{2/10}$ & $Acc_{3/10}$ & $Acc_{4/10}$ & $\overline{SRCC}$& $\overline{PCC}$  \\\hline
Baseline\_L                & 24.5 & -- & --   & --   & 41.0   & --   & -- & --   & --   & --    \\
A2-RL \cite{li2018a2}      & 23.0 &--  & --   & --   & 38.5   & --   & -- & --   & --   & --    \\
VPN\cite{wei2018good}      & 40.0 &-- & -- & -- & 49.5 & --   & -- & --   & --   & --    \\
VFN\cite{chen2017learning} & 27.0 &28.0 &27.2 &24.6 &39.0 &39.3 &39.0 &37.3 & 0.450 &0.470  \\
VEN\cite{wei2018good}      & 40.5 &36.5 &36.7 &36.8 &54.0 &51.0 &50.4 &48.4 & 0.621 &0.653  \\
GAIC (Conf.)\cite{zeng2019reliable}  &53.5 &51.5 &49.3 &46.6 &71.5 &70.0 &67.0 & 65.5 & 0.735 &0.762    \\\hline
GAIC (Mobile-V2)           & \textbf{62.5} & \textbf{58.3} & \textbf{55.3} & \textbf{52.5} & \textbf{78.5} & \textbf{76.2} & \textbf{74.8} & \textbf{72.3} & \textbf{0.783} & \textbf{0.806}   \\
GAIC (Shuffle-V2)          & 61.5 & 56.8 & 54.8 & 52.0 & \textbf{78.5} & 75.5 & 73.8 & 71.3 & 0.774 & 0.801   \\\hline\hline
& & & & & & & & & & \\[-1em]
Method                     & $Acc^w_{1/5}$ & $Acc^w_{2/5}$ & $Acc^w_{3/5}$ & $Acc^w_{4/5}$ & $Acc^w_{1/10}$ & $Acc^w_{2/10}$ & $Acc^w_{3/10}$ & $Acc^w_{4/10}$  & FPS (GPU)$^\star$ &FPS (CPU) \\\hline
Baseline\_L                & 15.6   & -- & --   & --   & 26.9   & --   & -- & --      & --  & -- \\
A2-RL \cite{li2018a2}      & 15.3   & -- & --   & --   & 25.6   & --   & -- & --      & 5  & 0.05 \\
VPN\cite{wei2018good}      & 19.5   & -- & --   & --   & 29.0   & --   & -- & --      & 75  & 0.8 \\
VFN\cite{chen2017learning} & 16.8 & 13.6 & 12.5 & 11.1 & 25.9 & 22.1 & 20.7 & 19.1 & 0.5  & 0.005\\
VEN\cite{wei2018good}      & 20.0 & 16.1 & 14.2 & 12.8 & 30.0 & 25.9 & 24.2 & 23.8 & 0.2  & 0.002\\
GAIC (Conf.)\cite{zeng2019reliable}  & 37.6 & 33.9 & 31.5 & 30.0 &53.7 &49.4 & 48.4 & 46.9 & 125  & 1.2  \\\hline
GAIC (Mobile-V2)           & 39.6 & 39.1 & 38.3 & 36.2 & 56.9 & \textbf{56.5} & \textbf{55.9} & \textbf{54.4} & \textbf{200}  & 6  \\
GAIC (Shuffle-V2)          & \textbf{40.3} & \textbf{39.4} & \textbf{38.6} & \textbf{37.2} & \textbf{57.3} & 55.0 & 54.7 & 53.6 & 142 &  \textbf{12}  \\\hline
\end{tabular}
\end{table*}

\subsubsection{Comparison methods}

Though a number of image cropping methods have been developed \cite{wang2017deep,deng2017image,chen2017quantitative,chen2017learning,deng2017aesthetic,guo2017automatic,li2018a2,wei2018good}, many of them do not release the source code or executable program. We thus compare our method, namely Grid Anchor based Image Cropping (GAIC), with the following baseline and recently developed state-of-the-art methods whose source codes are available.

\textbf{Baseline\_L:} The baseline\_L does not need any training. It simply outputs the largest crop among all eligible candidates. The result is similar to the ``baseline\_N" mentioned in Table \ref{table:Performance comparison}, i.e., the source image without cropping.

\textbf{VFN \cite{chen2017learning}:} The View Finding Network (VFN) is trained in a pair-wise ranking manner using professional photographs crawled from the Flickr. High-quality photos were first manually selected, and a set of crops were then generated from each image. The ranking pairs were constructed by always assuming that the source image has better quality than the generated crops.

\textbf{VEN and VPN \cite{wei2018good}:} Compared with VFN, the View Evaluation Network (VEN) employs more reliable ranking pairs to train the model. Specifically, the authors annotated more than 1 million ranking pairs using a two-stage annotation strategy. A more efficient View Proposal Network (VPN) was proposed in the same work, and it was trained using the predictions of VEN. The VPN is based on the detection model SSD \cite{liu2016ssd}, and it outputs a prediction vector for 895 predefined boxes.

\textbf{A2-RL \cite{li2018a2}:} The A2RL is trained in an iterative optimization manner. The model adjusts the cropping window and calculates a reward (based on predicted aesthetic score) for each step. The iteration stops when the accumulated reward satisfies some termination criteria.

\subsubsection{Qualitative comparison}

To demonstrate the advantages of our cropping method over previous ones, we first conduct qualitative comparison of different methods on various scenes including single object, multi-objects, building and landscape. Note that these images are out of any existing cropping databases.
In the first set of comparison, we compare all methods under the setting of returning only one best crop. Each model uses its default candidate crops generated by its source code except for VFN, which does not provide such code and uses the same candidates as our method. The results are shown in Fig. \ref{figure:qua comp top1}. We can make several interesting observations. Both VFN and A2-RL fail to robustly remove distracting elements in images. VFN sometimes cuts out important content, while A2-RL simply returns the source image in many cases. VEN and our GAIC model can stably output visually pleasing crops. The major differences lie in that VEN prefers more close-up crops while our GAIC tends to preserve as much useful information as possible.

A flexible cropping system should be able to output acceptable results under different requirements in practice, e.g., different aspect ratios. In this case, we generate multi-scale candidate crops with fixed aspect ratio and feed the same candidate crops into each of the competing models. In Fig. \ref{figure:qua comp fixed}, we show the top-1 returned crops by the competing methods under three most commonly used aspect ratios: 16:9, 4:3 and 1:1.  The A2-RL is not included because it does not support this test. Again, our model outputs the most visually pleasing crops in most cases.

\begin{figure*}[t]
\centering
\subfigure{
\begin{minipage}[b]{1.0\linewidth}
\centering
\includegraphics[width=1.0\textwidth]{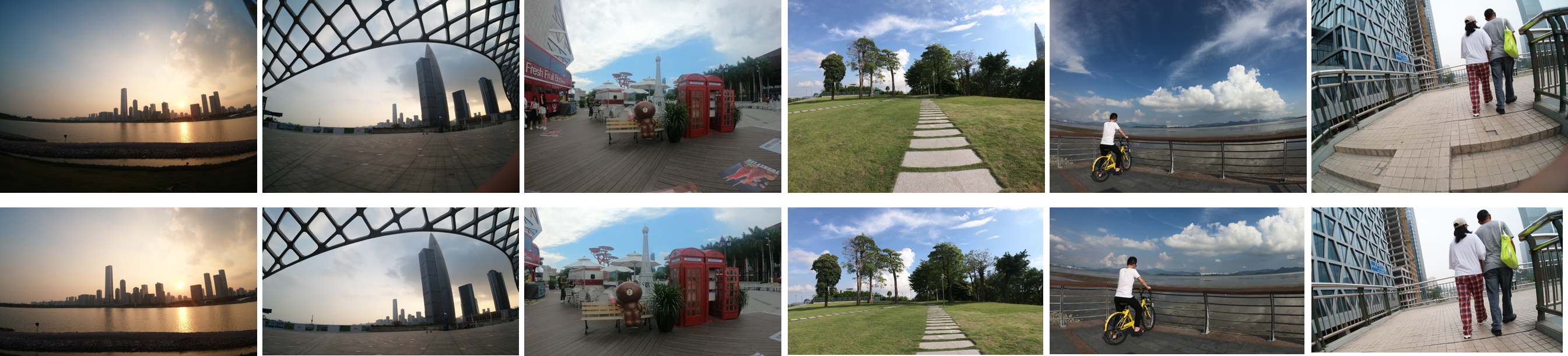}
\end{minipage}}%
\caption{16:9 crops generated by our model on images taken by wide lens action cameras. First row: 4:3 raw images captured by action cameras. Second row: 16:9 images generated by our model.}
\label{figure:widelens}
\end{figure*}

\subsubsection{Quantitative comparison}

We then perform quantitative comparisons by using the metrics defined in Section \ref{Evaluation metrics}.  Among the competitors, VFN, VEN and our GAIC support predicting scores for all the candidate crops provided by our database, thus they can be quantitatively evaluated by all the defined evaluation metrics. VPN uses its own pre-defined cropping boxes which are different from our database, and Baseline\_L and A2-RL output only one single crop. Therefore, we can only calculate $Acc_{1/5}$, $Acc^w_{1/5}$, $Acc_{1/10}$ and $Acc^w_{1/10}$ for them. We approximate the output boxes by VPN and A2-RL to the nearest anchor box in our database when calculating these accuracy indexes. The results of all competing methods on all the defined metrics are shown in Table \ref{table:others}.

We can draw several conclusions from the quantitative results. First, one can see that both A2-RL and VFN only obtain comparable performance to Baseline\_L. This is mainly because A2-RL is supervised by a general aesthetic classifier in training while the general aesthetic supervision across images cannot accurately discriminate different crops within one image, and the ranking pairs used in VFN are not very reliable because crops generated by well-composed images do not necessarily have worse composition than the source image. Although using the same pairwise learning strategy, VEN obtains much better performance than VFN by collecting more reliable ranking pairs through human annotations, which proves the necessity of human annotations for the cropping task. VPN performs slightly worse than VEN as expected because it is supervised by the predictions of VEN. Our model in the conference version already outperforms VEN by a large margin on all the evaluation metrics, benefitting from our dense annotated dataset which provides richer supervised information compared to the pair-wise ranking annotations used by VEN. Employing more efficient CNN architectures and more effective multi-scale features, our new models further significantly boost the cropping performance than our conference version on all the metrics.

\subsubsection{Running speed comparison}

A practical image cropping model should also have fast speed for real-time implementation. In the last two columns of Table \ref{table:others}, we compare the running speed in terms of frame-per-second (FPS) on both GPU and CPU for all competing methods. All models are tested on the same PC with i7-6800K CPU, 64G RAM and one GTX 1080Ti GPU, and our method is implemented under the PyTorch toolbox. As can be seen, our GAIC model based on the MobileNetV2 runs at 200 FPS on GPU and 6 FPS on CPU, and its counterpart based on the ShuffleNetV2 runs at 142 FPS on GPU and 12 FPS on CPU, both of which are much faster than the other competitors. It is worth mentioning that the GPU speeds in our testing are inconsistent with the CPU speeds because some operations such as group convolution and channel shuffle in the MobileNetV2 and ShuffleNetV2 are not well supported in PyTorch to make full use of the GPU computational capability.
The other models are much slower because they either employ heavy CNN architectures (VPN, GAIC (Conf.)), or need to individually process each crop (VFN and VEN) or need to iteratively update the cropping window several times (A2-RL), making them hard to be used in practical applications with real-time implementation requirement.

\subsubsection{Results on previous datasets}

As discussed in the introduction section, the limitations of previous image cropping databases and evaluation metrics make them unable to reliably reflect the cropping performance of a method. Nonetheless, we still evaluated our model on the ICDB \cite{yan2013learning} and FCDB \cite{chen2017quantitative} using the IoU as metric for reference of interested readers.
Since some groundtruth crops on these two databases have uncommon aspect ratios, we did not employ the aspect ratio constraint when generating candidate crops on these two datasets. We found that the value of $\lambda$ defined in the area constraint (Eq. \ref{equ:area constraint}) largely affects the performance of our model on the ICDB. We tuned $\lambda$ for the MobileNetV2 and ShuffleNetV2 based models and report their best results in Table \ref{table:Performance comparison}. However, like most previous methods, our models still obtain even smaller IoU than the baselines on the ICDB dataset and slightly better result on the FCDB dataset.
In contrast, as shown in previous subsections, a well trained model on our GAICD can obtain much better performance than the baseline. These results further prove the advantages of our new database as well as the associated metrics compared to previous ones.

\subsection{Application to action cameras}

We also evaluate the generalization capability of our model on a practical application: automatically cropping the images captured by action cameras. The action cameras usually have wide lens for capturing large field of view which is inevitably associated with severe lens distortion. We tested our trained model on 4:3 images taken by GoPro Hero 7 and DJI Osmo Pocket, and generated 16:9 crops. The results on six scenes are shown in Fig. \ref{figure:widelens}. We found that the model trained on our dataset can generalize well to the images with large lens distortion, because the lens distortion does not severely change the spatial arrangement of image content.

\section{Conclusion and Discussion}

We analyzed the limitations of existing formulation and databases on image cropping, and proposed a more reliable and efficient formulation for practical image cropping, namely grid anchor based image cropping (GAIC). A new benchmark was constructed, which contains 1,236 source images and 106,860 annotated crops. Three new types of metrics were defined to reliably and comprehensively evaluate the cropping performance on our database. We also designed very lightweight and effective cropping models by considering the special properties of cropping. Our GAIC model can robustly output visually pleasing crops under different aspect ratios. It runs at a speed up to 200 FPS on one GTX1080Ti GPU and 12 FPS on CPU, enabling real-time implementations on mobile devices.

There remain some limitations in our work, which leave much space for improvement. Firstly, our GAIC dataset is still limited in size considering the billions of photos generated in each day. It is expected that larger scale cropping dataset with reliable annotations can be constructed in the future. Second, the accuracies especially the rank weighted accuracies need further improvement. The cropping models are expected to learn more discriminative representations of photo composition in order to more accurately return the best crops.


%

\bibliographystyle{IEEETran}
\bibliography{egbib}

\ifCLASSOPTIONcaptionsoff
  \newpage
\fi

\end{document}